\pdfoutput=1

\documentclass[11pt]{article}
\usepackage{authblk}
\PassOptionsToPackage{hyphens}{url}
\usepackage[]{acl}
\usepackage{times}
\usepackage{latexsym}
\usepackage[size=tiny]{todonotes}
\usepackage{enumitem}
\usepackage{amsmath}
\usepackage{booktabs}
\usepackage{graphicx}
\usepackage[normalem]{ulem}
\useunder{\uline}{\ul}{}
\usepackage{multirow}
\usepackage{subfigure}
\usepackage{float}
\usepackage{stfloats}
\usepackage{bbding}
\newcommand{\Rmnum}[1]{\uppercase\expandafter{\romannumeral #1\relax}}

\newcommand{\se}[1]{\textcolor{black}{#1}}
\newcommand{\yc}[1]{\textcolor{black}{#1}}
\newcommand{\final}[1]{\textcolor{black}{#1}}
\newcommand{\sen}[1]{\textcolor{black}{#1}}

\usepackage[T1]{fontenc}

\usepackage[utf8]{inputenc}

\usepackage{microtype}

\title{Reproducibility Issues for BERT-based Evaluation Metrics} 

\author[1]{Yanran Chen}
\author[2]{Jonas Belouadi}
\author[2]{Steffen Eger}
\affil[1]{Computer Science Department, Technical University of Darmstadt, Germany}
\affil[2]{NLLG, Faculty of Technology, Bielefeld University, Germany}
\affil[1]{\texttt{yanran.chen@stud.tu-darmstadt.de}}
\affil[2]{\texttt {\{jonas.belouadi,steffen.eger\}@uni-bielefeld.de}}

\begin{document}
\maketitle
\begin{abstract}
Reproducibility is of utmost concern in machine learning and natural language processing (NLP). In the field of natural language generation (especially machine translation), the seminal paper of \citet{post2018call} has pointed out problems of reproducibility of the dominant metric, BLEU, at the time of publication. Nowadays, BERT-based evaluation metrics considerably outperform BLEU. In this paper, we ask whether results and claims from four recent BERT-based metrics can be reproduced. We find that reproduction of claims and results often fails because of (i) heavy undocumented preprocessing involved in the metrics, (ii) missing code and (iii) reporting weaker results for the baseline metrics. (iv) In one case, the problem stems from correlating not to human scores but to a wrong column in the csv file, inflating scores by 5 points. Motivated by the impact of preprocessing, we then conduct a second study where we examine its effects more closely (for one of the metrics). We find that preprocessing can have large effects, especially for highly inflectional languages. In this case, the effect of preprocessing may be larger than the effect of the aggregation mechanism (e.g., greedy alignment vs.\ Word Mover Distance). 
\end{abstract}

\section{Introduction}

Reproducibility is a core aspect in machine learning (ML) and natural language processing (NLP). It requires that claims and results of previous work can independently be reproduced and is a prerequisite to trustworthiness. 
The last few years have seen vivid interest in the topic and many issues of non-reproducibility have been pointed out, leading to claims of a ``reproducibility crisis'' in science \citep{baker2016reproducibility}. 
In the field of evaluation metrics for natural language generation (NLG),
the seminal work of \citet{post2018call} 
has demonstrated how different preprocessing schemes  
can lead to substantially different results when using the dominant metric at the time, BLEU \citep{Papineni2002BleuAM}. 
Thus, when researchers employ such different preprocessing steps (a seemingly innocuous decision), 
this can directly lead to reproducibility failures of (conclusions 
regarding)
metric performances. 

Even though BLEU and similar lexical-overlap metrics still appear to dominate the landscape of NLG (particular MT) evaluation \citep{marie-etal-2021-scientific}, it is obvious that  
metrics which measure surface level overlap are unsuitable for evaluation, especially for modern text generation systems with better paraphrasing capabilities \citep{mathur-etal-2020-tangled}. 
As a remedy, multiple higher-quality metrics based on BERT and its extensions have been proposed in the last few years \citep{zhang2019bertscore,zhao2019moverscore}. 
In this work, we  
investigate whether these more recent metrics have better reproducibility properties, thus filling a gap for the newer paradigm of metrics. We have good reason to suspect that reproducibility will be better: (i) 
as a response to the identified problems, recent years have seen  
many efforts in the NLP and ML communities to improve reproducibility, e.g., by requiring authors to fill out specific check lists.\footnote{E.g., \url{https://aclrollingreview.org/responsibleNLPresearch/}.} 
  (ii) Designers of novel evaluation metrics should be particularly aware of reproducibility issues, as reproducibility is a core concept of proper evaluation of NLP models \citep{eval4nlp-2021-evaluation}. 
 
 Our results are disillusioning: out of four metrics we tested, three exhibit (severe) reproducibility issues. The problems relate to (i) heavy use of (undocumented) preprocessing, (ii) missing code, (iii) reporting lower results for competitors,  
 and (iv) correlating with the wrong columns in the evaluation csv file. 
 Motivated by the findings on the role of preprocessing and following \citet{post2018call}, we then study its impact more closely in the second part of the paper (for those metrics making use of it), finding that it can indeed lead to substantial performance differences also for BERT-based metrics. 
 \final{The code for this work is available at {\small \url{https://github.com/cyr19/Reproducibility}}. }

\section{Related Work}
Relevant prior work to this work includes BERT-based evaluation metrics (Section \ref{sec:rela1}) and reproducibility in NLP (Section \ref{sec:rela2}).

\subsection{BERT-based Evaluation Metrics}
\label{sec:rela1}
In recent years,  
many strong automatic evaluation metrics based on BERT \citep{devlin2018bert} or its variants have been proposed. It has been shown that those BERT-based evaluation metrics  
correlate much better with human judgements than traditional evaluation metrics such as BLEU \citep{Papineni2002BleuAM}. 
Popular supervised BERT-based evaluation metrics include BLEURT \citep{sellam-etal-2020-bleurt} and COMET \citep{rei2020comet}, which are trained on segment-level human judgements such as DA scores in WMT datasets.  
Unsupervised BERT-based evaluation metrics such as BERTScore \citep{zhang2019bertscore}, MoverScore \citep{zhao2019moverscore}, BaryScore \citep{colombo2021automatic} and XMoverScore \citep{zhao2020limitations} do not use training signals, thus potentially may generalize better to unseen language pairs \citep{Belouadi2022USCOREAE}. 
MoverScore, BaryScore, and BERTScore are \emph{reference-based} evaluation metrics. 
In contrast, \emph{reference-free} evaluation metrics directly compare system outputs to source texts. For MT, popular such metrics are Yisi-2 \citep{lo-2019-yisi}, XMoverScore, and SentSim \citep{song-etal-2021-sentsim}.

\subsection{Reproducibility in NLP}
\label{sec:rela2}

\citet{Cohen2018ThreeDO} define \textbf{replicability} as the ability to repeat the process of experiments and \textbf{reproducibility} as the ability to obtain the same results.  
They further categorize reproducibility along 3 dimensions: (1) reproducibility of a \emph{conclusion}, (2) reproducibility of a \emph{finding}, and (3) reproducibility of a \emph{value}. In a more recent study, \citet{belz-etal-2021-systematic} categorize reproduction studies according to the \emph{condition of the reproduction experiment}: (1) \emph{reproduction under the same condition}, i.e., re-using as \sen{similar} as possible resources and mimicking the authors' experimental procedure as closely as possible; (2) \emph{reproduction under varied conditions}, aiming to test whether the proposed methods can obtain similar results with some changes in the settings; (3) \emph{multi-test and multi-lab studies}, i.e., reproducing multiple papers using uniform methods and multiple teams attempting to reproduce the same paper, respectively. 

In the first part of this work,  
our reproductions follow 
the first type 
described by 
\citet{belz-etal-2021-systematic},  
i.e., we adhere to the original experimental setup and re-use the resources provided by the authors whenever possible, aiming at exact reproduction. 
The second part 
falls into the second category of reproduction study described by \citet{belz-etal-2021-systematic}, i.e., to change some settings on purpose to see if comparable results can be obtained. 
 
According to \citet{Fokkens2013OffspringFR} and \citet{wieling-etal-2018-squib}, the main challenge \sen{for reproducibility} is the unavailability of the source code and data. 
\citet{Dakota2017TowardsRI} study reproducibility for text mining. They show that 80\% of the failed reproduction attempts were due to the lack of information about the datasets.  
To investigate  
the availability of source data, \citet{Mieskes2017AQS} conducted quantitative analyses on the publications from five conferences. 
They \sen{found} that though 40\% of the papers reported having collected or changed existing data, only 65.2\% of them provided the links to download the data; 18\% of them were invalid. 
Similarly, \citet{wieling-etal-2018-squib} assessed the availability of both source code and data of papers 
from two ACL conferences (2011 and 2016).  
When comparing 2016 to 2011, the availability of both data and code increased, suggesting a growing trend of sharing resources for reproduction. However, even using the same code and data, they could only recreate identical values for one paper. 
More recently, \citet{belz-etal-2021-systematic} analyzed 34 reproduction studies under the same condition (re-using the original resources when possible) for NLP papers. They found that only a small portion (14.03\%) of values could be exactly reproduced and the majority (59.2\%) of the reproduced values lead to worse results. Moreover, 1/4 deviations are $>$5\%.

In NLG, 
\citet{post2018call} attests to the non-comparability of BLEU \citep{Papineni2002BleuAM} scores across different papers. He argues that there are four causes. \textbf{First}, BLEU is a parameterized approach; 
he shows that on WMT17 \citep{wmt17}, the BLEU score for en-fi, 
increases by roughly 3\%  
Pearson 
from changing parameters regarding multiple references.
The \textbf{second} issue, which is regarded as the most critical,  
is the use of different preprocessing schemes. Among these, 
tokenization of the references plays a key role.
The \textbf{third} problem is that preprocessing details are 
often omitted in papers.  
The \textbf{fourth} problem is different versions of datasets, in his case a particular problem with the en-de language pair in WMT14 \citep{machacek-bojar-2014-results}. The reproducibility issue of BLEU has also been verified by \citet{belz2022quantified}, using their novel approach, which is designed to quantify the degree of reproducibility.

\section{Datasets \& Metrics}
In our reproduction experiments (Section \ref{sec:reproduction}), following \citet{zhang2019bertscore}, \citet{zhao2019moverscore} and \citet{colombo2021automatic}, we use WMT15-18 \citep{wmt15,wmt16,wmt17,wmt18} for MT evaluation. Besides, 
we follow \citet{zhao2019moverscore} to use 
TAC2008\footnote{\url{https://tac.nist.gov/2008/}} and TAC2009\footnote{\url{https://tac.nist.gov/2009/}}  
for text summarization evaluation, MSCOCO \citep{guo-etal-2018-meteor} for image captioning (IC) evaluation, and BAGEL \citep{wen2015semantically} and SFHOTEL \citep{mairesse2010phrase} for data-to-text generation (D2T) evaluation. For the reference-free metric SentSim, we will mainly report results on the MLQE-PE dataset \citep{fomicheva2020mlqe}. 
In further experiments (Section \ref{sec:sensitivity}), we consider  
WMT19 \citep{wmt19} for MT as well.  
The datasets for each NLG task are described in detail in the appendix (Section \ref{sec:dataset_description}). 
For our reproduction attempts, we consider MoverScore, 
BERTScore,  
BaryScore, and SentSim.

\paragraph{Metrics}
\textbf{MoverScore}
measures semantic similarity between reference and hypothesis by aligning semantically similar words and computing the distance 
between these words 
using the Word Mover Distance \citep{kusner2015word}. 
%
\textbf{BERTScore}
calculates the cosine similarity \se{(of BERT representations)} for each token in the reference  
with each token in the hypothesis,   
and uses greedy alignment to obtain  
the similarity scores between sentences. It has three variants: Recall, Precision, and F1. 
%
\textbf{BaryScore}
computes the Wasserstein distance (i.e., Earth Mover Distance \citep{rubner2000earth}) between the barycentric distribution \citep{agueh2011barycenters} of the contexual representations of reference and hypothesis to measure the dissimilarity between them.
%
\textbf{SentSim} \final{has both reference-free and -based versions; we experiment with its reference-free version in this work, which}
combines sentence- (based on \citet{reimers-gurevych-2020-making}) and word-level models (extending a.o.\ BERTScore to the multilingual case) to score a pair of source text and hypothesis. 
\section{Reproduction Attempts}
\label{sec:reproduction}
Our main focus will be 
to reproduce the results on machine translation (\textbf{MT})  
reported in \citet{zhang2019bertscore}, \citet{zhao2019moverscore},  \citet{colombo2021automatic} and \citet{song-etal-2021-sentsim}. 
\subsection{Reproduction on MT}
 
\sen{At first, we 
examine the three reference-based metrics.} 
MoverScore, BaryScore and BERTScore  
were \sen{all} originally evaluated on MT but with different WMT datasets \citep{wmt15,wmt16,wmt17,wmt18}. 
\citet{zhang2019bertscore} used WMT18 \citep{wmt18} as the main evaluation dataset. 
\citet{zhao2019moverscore}  
reported the results on WMT17 \citep{wmt17} for both MoverScore and BERTScore-F1. \citet{colombo2021automatic} compared their metric BaryScore with MoverScore and BERTScore-F1 on WMT15 \citep{wmt15} and WMT16 \citep{wmt16}. 
MoverScore claims to outperform BERTScore (which was published earlier on Arxiv), and BaryScore claims to outperform the earlier two. 

\begin{table*}[!htb]
\centering
\resizebox{0.75\textwidth}{!}{\small %
\begin{tabular}{@{}c|lcccccccc@{}}
\toprule
                            & metric       & cs-en & de-en & et-en & fi-en          & ru-en & tr-en & zh-en & avg   \\ \midrule
\multirow{3}{*}{Reproduced} 
& BaryScore-W  & 0.360 & 0.525 & 0.379 & 0.280       & 0.322 & 0.254 & \textbf{0.252} & 0.339 \\
& MoverScore-1 & 0.362 & 0.529 & 0.391 & \textbf{0.297} & 0.338 & 0.288 & 0.244 & 0.350 \\

& BERTScore-F1 & {\color[HTML]{009901} \textbf{0.376}} & {\color[HTML]{009901} \textbf{0.538}} &  \textbf{0.393} & {\color[HTML]{009901} 0.295} & {\color[HTML]{009901} \textbf{0.341}} & {\color[HTML]{009901} \textbf{0.290}} & {\color[HTML]{009901} 0.244} & {\color[HTML]{009901} \textbf{0.354}} \\ \midrule
Reported                    & BERTScore-F1 & 0.375 & 0.535 & 0.393 & 0.294          & 0.339 & 0.289 & 0.243 & 0.353 \\ \bottomrule
\end{tabular}%
}
\caption{Reproduction: Segment-level Kendall's $\tau$ on WMT18 to-English language pairs using the evaluation script provided by \citet{zhang2019bertscore}. Reported values are taken from \citet{zhang2019bertscore}. Values in green denote  reproduced results that are better than the reported. Bold values refer to the best reproduced results with the BERT-base-uncased model.}
\label{tab: reproduction wmt18}

\centering
\resizebox{0.8\textwidth}{!}{ 
\begin{tabular}{@{}c|lllllllll@{}}
\toprule
 &
  metric &
  cs-en &
  de-en &
  fi-en &
  lv-en &
  ru-en &
  tr-en &
  zh-en &
  avg \\ \midrule
\multirow{4}{*}{Reproduced} &
  BaryScore-W &
  0.646 &
  0.652 &
  0.819 &
  0.689 &
  0.697 &
  \textbf{0.737} &
  0.719 &
  0.709 \\
 &
  MoverScore-1 &
  \textbf{0.660} &
  \textbf{0.690} &
  0.806 &
  0.685 &
  \textbf{0.736} &
  0.732 &
  \textbf{0.720} &
  \textbf{0.718} \\
 &
  BERTScore-F1 &
  {\color[HTML]{FE0000} 0.655} &
  {\color[HTML]{FE0000} 0.682} &
  {\color[HTML]{009901} \textbf{0.823}*} &
  {\color[HTML]{009901} \textbf{0.713}} &
  {\color[HTML]{FE0000} 0.725} &
  {\color[HTML]{009901} 0.718} &
  {\color[HTML]{009901} 0.712} &
  {\color[HTML]{FE0000} 0.718} \\
 &
  MoverScore-1$^{+}$ &
  0.670* &
  0.708* &
  {\color[HTML]{FE0000} 0.821} &
  {\color[HTML]{FE0000}0.717*} &
  0.738* &
  0.762* &
  0.744* &
  {\color[HTML]{FE0000}0.737*} \\ \midrule
\multirow{2}{*}{Reported} & MoverScore-1$^{+}$ & 0.670       & 0.708*       & 0.835* & 0.746* & 0.738*       & 0.762*       & 0.744*       & 0.743* \\
                          & BERTScore-F1  & 0.670 &  0.686 & 0.820 & 0.710 & 0.729 &  0.714 &  0.704 &  0.719 \\ \bottomrule
\end{tabular}%
}
\caption{Reproduction: Segment-level Pearson's $r$ on WMT17 to-English language pairs using evaluation script provided by \citet{zhao2019moverscore}. Reported results are cited from \citet{zhao2019moverscore}. $^{+}$ refers to using the finetuned BERT-based-uncased model on MNLI. Values in green/red denote the reproduced results are better/worse than the reported. Bold values refer to the best results with BERT-base-uncased model. Values with * denote the best reproduced/reported results.}
\label{tab: reproduction wmt17}

\end{table*}

We evaluate the three metrics with the same BERT model (BERT-base-uncased) 
on all MT datasets mentioned above, using the reproduction resources provided by the authors of each metric. 
We also evaluate MoverScore and BaryScore on a BERT model finetuned on NLI \citep{wang-etal-2018-glue} (as in the original papers). 
The code and data for reproduction were released on their respective githubs.\footnote{BERTScore (WMT18): \url{https://github.com/Tiiiger/bert_score/tree/master/reproduce}; MoverScore (WMT17): \url{https://github.com/AIPHES/emnlp19-moverscore/tree/master/examples}; BaryScore (WMT15-16): \url{https://github.com/PierreColombo/nlg_eval_via_simi_measures/tree/main/raw_score}.} 
In our reproduction experiments, we use the metrics with the configurations 
found in their evaluation scripts or papers. 
Although \citet{zhao2019moverscore} also reported the results for BERTScore-F1, they did not provide information about the used parameter settings. 
Similarly, \citet{colombo2021automatic} evaluated the other two metrics on WMT15-16, but except for the model choice, all other settings are unclear. Moreover, except for \citet{zhang2019bertscore}, who explicitly state which results were obtained using IDF-weighting, the authors of the other two approaches did not mention this in their papers. 
For unclear metric configurations, we keep them \sen{at} 
default. 
The configurations used here are:

\begin{itemize}[topsep=2pt,itemsep=-1pt,leftmargin=*]
    \item \textbf{BERTScore}
    We report the reproduced results for BERTScore-F1 that uses 
    BERT-base-uncased, with the default layer 9 of the BERT representation for this model, and IDF-weighting.
    
    \item \textbf{MoverScore}
     We report the reproduced results for unigram MoverScore (MoverScore-1) using BERT-base-uncased or its finetuned version on MNLI, the last five layers from BERT aggregated by power means \citep{rueckle-etal-2018-pmeans}, IDF-weighting, punctuation removal and subwords removal (only keep the first subword in each word).
     
    \item \textbf{BaryScore}
    We report the reproduced results for BaryScore\footnote{BaryScore outputs scores relying on Wasserstein distance and those relying on Sinkhorn distance \citep{cuturi2013sinkhorn} together. We report the results for Wasserstein distance, which are denoted as BaryScore-W or Bary-W.} that makes use of BERT-base-uncased or its finetuned version on MNLI\footnote{\final{As the authors of BaryScore did not release their finetuned model, we use the NLI model released by the authors of MoverScore for BaryScore.}}, the last five layers aggregated using Wasserstein Barycenter, and IDF-weighting.
    
\end{itemize}
The metrics with finetuned models are marked with $^{+}$ in the following.

\begin{table*}[]
\centering
\resizebox{0.95\textwidth}{!}{%
\begin{tabular}{@{}clllllllllllll@{}}
\toprule
\multicolumn{1}{l|}{} &
  \multicolumn{6}{c|}{WMT15} &
  \multicolumn{7}{c}{WMT16} \\
\multicolumn{1}{l|}{} &
  metric &
  \multicolumn{1}{c}{cs-en} &
  \multicolumn{1}{c}{de-en} &
  \multicolumn{1}{c}{fi-en} &
  \multicolumn{1}{c}{ru-en} &
  \multicolumn{1}{c|}{avg} &
  \multicolumn{1}{c}{cs-en} &
  \multicolumn{1}{c}{de-en} &
  \multicolumn{1}{c}{ru-en} &
  \multicolumn{1}{c}{fi-en} &
  \multicolumn{1}{c}{ro-en} &
  \multicolumn{1}{c}{tr-en} &
  \multicolumn{1}{c}{avg} \\ \midrule
\multicolumn{1}{c|}{} &
  BERT-F &
  {\color[HTML]{009901} 0.750} &
  {\color[HTML]{009901} \textbf{0.733}} &
  {\color[HTML]{009901} 0.752} &
  {\color[HTML]{009901} \textbf{0.745}} &
  \multicolumn{1}{l|}{{\color[HTML]{009901} 0.745}} &
  {\color[HTML]{009901} \textbf{0.747}} &
  {\color[HTML]{FE0000} 0.640} &
  {\color[HTML]{009901} 0.672} &
  {\color[HTML]{009901} 0.661} &
  {\color[HTML]{009901} \textbf{0.723}} &
  {\color[HTML]{FE0000} 0.688} &
  {\color[HTML]{009901} 0.689} \\
  \multicolumn{1}{c|}{} 
 &
  Mover-1 &
  {\color[HTML]{009901} 0.734} &
  {\color[HTML]{009901} 0.731} &
  {\color[HTML]{009901} 0.743} &
  {\color[HTML]{009901} 0.731} &
  \multicolumn{1}{l|}{{\color[HTML]{009901} 0.735}} &
  {\color[HTML]{009901} 0.740} &
  {\color[HTML]{009901} 0.633} &
  {\color[HTML]{009901} \textbf{0.676}} &
  {\color[HTML]{009901} 0.655} &
  {\color[HTML]{009901} 0.714} &
  {\color[HTML]{009901} 0.693} &
  {\color[HTML]{009901} 0.685} \\
  \multicolumn{1}{c|}{} 
 &
  Bary-W &
  {\color[HTML]{009901} \textbf{0.751}} &
  {\color[HTML]{FE0000} 0.731} &
  {\color[HTML]{009901} \textbf{0.769}} &
  {\color[HTML]{009901} 0.740} &
  \multicolumn{1}{l|}{{\color[HTML]{009901} \textbf{0.748}}} &
  {\color[HTML]{FE0000} 0.735} &
  {\color[HTML]{009901} \textbf{0.672}} &
  {\color[HTML]{FE0000} 0.659} &
  {\color[HTML]{009901} \textbf{0.673}} &
  {\color[HTML]{FE0000} 0.715} &
  {\color[HTML]{009901} \textbf{0.709}} &
  {\color[HTML]{009901} \textbf{0.694}} \\
  \multicolumn{1}{c|}{} 
 &
  Mover-1$^{+}$ &
  {\color[HTML]{009901} 0.745} &
  {\color[HTML]{009901} 0.755*} &
  {\color[HTML]{009901} 0.774} &
  {\color[HTML]{009901} 0.765*} &
  \multicolumn{1}{l|}{{\color[HTML]{009901} 0.760}} &
  {\color[HTML]{009901} 0.765*} &
  {\color[HTML]{009901} 0.676} &
  {\color[HTML]{009901} 0.696*} &
  {\color[HTML]{009901} 0.707*} &
  {\color[HTML]{009901} 0.742*} &
  {\color[HTML]{009901} 0.736} &
  {\color[HTML]{009901} 0.720*} \\
\multicolumn{1}{c|}{\multirow{-5}{*}{Reproduced}} &
  Bary-W$^{+}$ &
  {\color[HTML]{FE0000} 0.753*} &
  {\color[HTML]{FE0000} 0.755} &
  {\color[HTML]{FE0000} 0.787*} &
  {\color[HTML]{FE0000} 0.763} &
  \multicolumn{1}{l|}{{\color[HTML]{FE0000} 0.764*}} &
  {\color[HTML]{FE0000} 0.758} &
  {\color[HTML]{009901} 0.700*} &
  {\color[HTML]{FE0000} 0.677} &
  {\color[HTML]{009901} 0.706} &
  {\color[HTML]{FE0000} 0.732} &
  {\color[HTML]{009901} 0.744*} &
  {\color[HTML]{FE0000} 0.720} \\ \midrule
  \multicolumn{1}{c|}{} 
 &
  BERT-F &
  \textbf{0.743} &
  0.722 &
  0.747 &
  \textbf{0.740} &
  \multicolumn{1}{l|}{0.738} &
  0.741 &
  \textbf{0.653} &
  0.651 &
  0.654 &
  0.702 &
  \textbf{0.707} &
  0.685 \\
  \multicolumn{1}{c|}{} 
 &
  Mover &
  0.688 &
  0.718 &
  0.700 &
  0.686 &
   \multicolumn{1}{l|}{0.698} &
  0.674 &
  0.609 &
  0.644 &
  0.631 &
  0.642 &
  0.661 &
  0.644 \\
  \multicolumn{1}{c|}{}
 &
  Bary &
  0.742 &
  \textbf{0.741} &
  \textbf{0.766} &
  0.737 &
   \multicolumn{1}{l|}{\textbf{0.747}} &
  \textbf{0.742} &
  0.646 &
  \textbf{0.675} &
  \textbf{0.671} &
  \textbf{0.725} &
  0.693 &
  \textbf{0.692} \\
  \multicolumn{1}{c|}{}
 &
  Mover$^{+}$ &
  0.710 &
  0.711 &
  0.722 &
  0.673 &
  \multicolumn{1}{l|}{0.704} &
  0.707 &
  0.624 &
  0.640 &
  0.645 &
  0.664 &
  0.663 &
  0.657 \\
\multicolumn{1}{c|}{\multirow{-5}{*}{Reported}} &
  Bary$^{+}$ &
  0.759* &
  0.758* &
  0.799* &
  0.776* &
  \multicolumn{1}{l|}{0.773*} &
  0.766* &
  0.685* &
  0.694* &
  0.702* &
  0.743* &
  0.738* &
  0.721* \\ \bottomrule
\end{tabular}%
}
\caption{Reproduction: Segment-level Pearson's r on WMT15-16 using evaluation script provided by \citet{colombo2021automatic}. Reported values are cited from \citet{colombo2021automatic}. $^{+}$ represents using the fine-tuned BERT-base-uncased model on MNLI. Values in green/red denote the reproduced results are better/worse than the reported. Bold values refer to the best results with BERT-base-uncased model. Values with * denote the best reproduced/reported results.}
\label{tab: reproduction wmt1516}
\end{table*}

\paragraph{Results}
As Table \ref{tab: reproduction wmt18} 
shows, we do not obtain identical results for BERTScore-F1 with \citet{zhang2019bertscore} on \textbf{WMT18} to-English language pairs. 
The maximal deviation between the reported and reproduced results can be seen on the evaluated data for 
de-en -- around 0.003 absolute Pearson's $r$.  
Most of the deviations are about 0.001. 
This might be because of tiny differences in rounding strategies 
\final{and random seeds\footnote{\final{We noted that this evaluation script produces different results every time, but the discrepancy across different runs in the averaged correlations is tiny ($\sim$0.001-0.002).}}} etc.
Further, among the three evaluation metrics, BERTScore-F1 performs best, whereas BaryScore is worst. 

Table \ref{tab: reproduction wmt17} displays the reproduction results on \textbf{WMT17} to-English language pairs, leveraging the resources from \citet{zhao2019moverscore}. As for MoverScore-1$^{+}$, 5 out of 7 values can be perfectly reproduced (excluding the average value). The unreproducible results on fi-en and lv-en are 0.012 and 0.031 lower than the reported,  
respectively. \se{On personal communication, the authors told us that they changed the preprocessing for these settings, which is impossible to identify from the released paper or code}. 
We obtain comparable average value for BERTScore-F1 with \citet{zhao2019moverscore} (0.718 vs.\ 0.719),  
but the results on individual language pairs differ. 
Except for fi-en, MoverScore-1$^{+}$ correlates better with humans than BERTScore-F1, which is in line with the observation from \citet{zhao2019moverscore}. When applying the same BERT model, BaryScore performs slightly worse than the other two metrics, 
except for tr-en. 

Table \ref{tab: reproduction wmt1516} shows the results of the reproduction attempts on \textbf{WMT15-16} based on the code and data provided by \citet{colombo2021automatic}. \citet{colombo2021automatic} reported Pearson, Spearman and Kendall correlation with human ratings;  
we relegate the reproduction results for Kendall and Spearman correlation, which are similar to those for Pearson correlation, to Section \ref{sec:repro_wmt15}.  
We are not able to reproduce identical values for any evaluation metric, even for  
BaryScore. However, the reproduced results for BaryScore and BaryScore$^{+}$ are comparable with the reported -- around 0.001 Pearson off the reported average values in 3 out of 4 cases. 
For BERTScore-F1, the reproduced average values are around 0.005 Pearson better than the reported, while for MoverScore/MoverScore$^{+}$, they are about 0.05 Pearson better. 
\citet{colombo2021automatic} observed that BaryScore$^{+}$ performs best on all language pairs in WMT15-16, which is inconsistent with our observation:
MoverScore-1$^{+}$ outperforms BaryScore$^{+}$ on half the language pairs in these two datasets.  
With BERT-base-uncased, 
BaryScore performs best among the three evaluation metrics on these two datasets, however --- it achieves the highest correlation on 6 out of 10 language pairs. 
\vspace{-0.2cm}
\paragraph{Summary}
We can rarely reconstruct identical values but  
obtained comparable results for the three discussed metrics, even when some of the metric configurations are missing.  
However, we can overall not reproduce the \emph{conclusions} for three main reasons: (i) authors report lower scores for competitor metrics;  
(ii) authors selectively evaluate on specific datasets (maybe omitting those for which their metrics do not perform well?); (iii) unlike the authors of BERTScore, the authors of BaryScore and MoverScore do not provide a unique hash,  
making reproduction of the original values more difficult; (iv) undocumented preprocessing involved.

Following the three reproduction attempts, we cannot conclude that the newer approaches are better than the prior ones (BertScore),  
as \citet{zhao2019moverscore} and \citet{colombo2021automatic} claim. 
\se{We also point out that the three metrics perform very similar when using the same underlying BERT model; using a BERT model fine-tuned on NLI seems to have a bigger impact. This casts some doubt on whether the more complicated word alignments (as used in BaryScore and MoverScore) really have a critical effect.}

\begin{table*}[!htb]
\centering
\resizebox{0.8\textwidth}{!}{%
\begin{tabular}{@{}c|lcccccccc@{}}
\toprule
                           & metric                   & en-de & en-zh & ru-en & ro-en & et-en & ne-en & si-en & avg   \\ \midrule
 &
  SentSim(BERTScore-based) &
  {\color[HTML]{FE0000} 6.15} &
  {\color[HTML]{FE0000} 22.23} &
  {\color[HTML]{FE0000} 47.30} &
  {\color[HTML]{009901} 78.55} &
  {\color[HTML]{FE0000} 55.09} &
  {\color[HTML]{009901} 57.09} &
  {\color[HTML]{009901} 51.14} &
  {\color[HTML]{FE0000} 45.36} \\
\multirow{-2}{*}{Fixed} &
  SentSim(WMD-based) &
  {\color[HTML]{FE0000} 3.86} &
  {\color[HTML]{FE0000} 22.62} &
  {\color[HTML]{FE0000} 47.46} &
  {\color[HTML]{009901} 77.72} &
  {\color[HTML]{FE0000} 54.60} &
  {\color[HTML]{009901} 57.00} &
  {\color[HTML]{009901} 49.79} &
  {\color[HTML]{FE0000} 44.72} \\ \midrule
                           & SentSim(BERTScore-based) & 48.40 & 42.70 & 47.50 & 72.70 & 55.30 & 39.20 & 42.60 & 49.80 \\
\multirow{-2}{*}{Reported} & SentSim(WMD-based)       & 47.20 & 42.70 & 47.60 &
  72.40 & 55.30 & 39.00 & 42.60 & 49.50 \\ \midrule
                            & D-TP      & 25.90 & 32.10 &  ---  & 69.30 & 64.20 & 55.80 & 46.00 & 48.90 \\
\multirow{-2}{*}{Baselines} & D-Lex-Sim & 17.20 & 31.30 &  ---  & 66.30 & 61.20 & 60.00 & 51.30 & 47.90 \\ \bottomrule
\end{tabular}%
}
  \caption{Correlations of SentSim on MLQE-PE with model log-likelihoods (Reported), as
  erroneously done in the official paper, and with human judgments (Fixed). The
  green and red highlighted results on human judgments indicate that they are
  better or worse than the corresponding results computed with log-likelihoods.
  We cite baseline scores from \citet{Fomicheva2020UnsupervisedQE}.}%
  \label{tab:repro_sentsim}
\end{table*}

\paragraph{SentSim} 
For reference-free evaluation, \citet{song-etal-2021-sentsim} use MLQE-PE as
their primary evaluation dataset.  
They compare SentSim to
so-called glass-box metrics which actively incorporate the
MT system under test into the scoring
process~\citep{Fomicheva2020UnsupervisedQE}. 

Using the original model configuration, we were able to exactly reproduce the
reported scores for all SentSim variants on MLQE-PE\@. However, we noticed that
the provided code for loading the dataset does not retrieve human judgments but
averaged log-likelihoods of the NMT model used to generate the hypotheses.
Since computing correlations with model log-likelihoods is not meaningful
and the z-standardized means of the human judgments that should have been used
instead are in an adjacent column of the dataset, we assume that this is an
off-by-one error. 

Table~\ref{tab:repro_sentsim} shows how much fixing this error affects the
achieved correlations of BERTScore- and WMD-based SentSim. The baselines were
not affected by this, as \citet{song-etal-2021-sentsim} copied their scores
from their original papers. Evaluation on human judgments leads to vast score
differences on many language pairs. This is especially noticeable for
English-German and English-Chinese language pairs, where the correlations
achieved with our fixed implementation are \sen{substantially} worse. This result is
much more in line with the findings of related research, which also notes very
poor performance for these languages on this
dataset~\citep{fomicheva2020mlqe,Specia2020FindingsOT}. \sen{We note that after fixing the error, SentSim falls below the baselines, which it had otherwise outperformed.}

\subsection{Reproduction for other tasks} 

\se{In Section \ref{sec:reproduction_other}, we reproduce results for other tasks, especially summarization, image captioning and data-to-text generation, with a focus on MoverScore. We find that we can only reproduce the reported results for summarization, 
and our results are on average 0.1 Pearson's $r$ (-12.8\%) down 
for IC and 0.06 Spearman's $\rho$ (-27.8\%) down for D2T generation. A reason is that the authors of MoverScore did not release their evaluation scripts and we can only speculate as to their employed preprocessing steps. As long as these are not reported in the original papers or released code, claims regarding performance of the metrics are hard to verify.\footnote{We cannot rule out the possibility that  
we made 
mistakes in our reproduction attempts (e.g., incorrect evaluation scripts or use of datasets), but the unavailability of the resources makes the detection of potential errors difficult.}}

\section{Sensitivity Analysis} 
\label{sec:sensitivity}
 
In the previous section, we have seen that preprocessing may play a vital role for obtaining state-of-the-art results (at least for some of the metrics). Similar to the case of BLEU \cite{post2018call}, we now examine this aspect in more detail. 

According to the papers and evaluation scripts,  
{MoverScore} uses the following main preprocessing steps (besides those handled by BERT):  
(i) \textbf{Subwords Removal}: discard BERT representations of all subwords except the first. (ii) \textbf{Punctuation Removal}: discard BERT representations of punctuations. (iii) \textbf{Stopwords Removal}: discard BERT representations of stopwords (only for summarization).\footnote{In Section \ref{sec:subwords}, we describe subword removal and stopword and punctuation removal used in MoverScore.} 
The preprocessing steps for {BERTScore} and {BaryScore} are only related to lowercasing and tokenization, both of which are handled by BERT. We observe that 
(i) MoverScore uses much more preprocessing than 
BERTScore and BaryScore 
on WMT datasets; 
(ii)  
authors 
may take different preprocessing steps for different tasks, e.g.,  
\citet{zhao2019moverscore}  
remove stopwords for summarization but not for MT. 

Besides preprocessing in a \sen{narrower} sense, all three considered evaluation metrics use \emph{parameters}.  
This makes them  
more flexible,  
but also complicates reproduction: the difference in one parameter setting can lead to reproduction failure.  
We study the impact of the parameters related to 
\textbf{IDF-weighting}. \sen{IDF-weighting} measures 
how critical a word is to a corpus; thus, it is corpus-dependent. The choice of corpus  
may lead to deviations of metric scores. 
    
MoverScore is the main experiment object in the remainder.  
Compared to  
the other metrics, 
its authors 
took more preprocessing steps to achieve the results in the\sen{ir} paper, suggesting that it is more likely to obtain uncomparable scores across different users when using MoverScore.
We will also investigate the sensitivity of  
BERTScore to the factors discussed above;  
we omit BaryScore and SentSim from further consideration. 
Importantly, we move beyond English-only evaluation, as reported in the original MoverScore paper. {This will estimate how much uncertainty there is from preprocessing when a user applies MoverScore to a non-English language pair, which requires new IDF corpora, new stopword lists and may have higher morphological complexity (which is related to choice of subwords).}

We use two statistics to quantify the sensitivity of the evaluation metrics. When there are only \textbf{two compared values} $a,b$, we compute  
Relative Difference (\texttt{RD}) to reflect the  
relative performance variation regarding a certain parameter. 
When there are \textbf{more than two compared values}, we compute 
Coefficient of Variation (\texttt{CV}) to reflect the extent of variability of the metric performance: 
    \begin{align*}
        & \texttt{RD}(a,b)=\frac{a-b}{b}*100\% 
,\\ & \texttt{CV}(x) = \frac{\sigma}{\mu}*100\% 
        \quad 
    \end{align*}
where $\sigma$ is the standard deviation and $\mu$ is the mean of a set of values $x$.   
Larger absolute values of the statistics indicate higher sensitivity of the evaluation metrics. 

We only consider MT and summarization evaluation in this part. 
In each experiment, we only adjust the settings of the tested factors and keep the others default (given in Section \ref{sec:default}). In addition to English \se{(``to-English'')}, we consider MT evaluation for other 6 languages \se{(``from-English'')}, \se{for which we use multilingual BERT}: Chinese (zh), Turkish (tr), Finnish (fi), Czech (cs), German (de), and Russian (ru). \sen{Note that in these cases, we compare a Chinese reference to a Chinese hypothesis and analogously for the other languages.}

\subsection{Stopwords Removal}
\label{stopwords set}
In this experiment, we consider 4 stopword settings including disabling stopwords removal and applying 3 different stopword lists for the examined languages. 
We obtain the stopword lists  
from the resources listed in  
Section \ref{sec:otherlangs}. 
We inspect the sensitivity of MoverScore-1, MoverScore-2 (MoverScore using bigrams) and BERTScore-F1 to stopword settings, despite that BERTScore does originally not  
employ stopwords. 

For English MT, we calculate \texttt{CV} of the correlations \se{with humans} over the 4 stopword settings for each language pair in the datasets, then average \texttt{CV}s over the language pairs in each dataset to obtain the average \texttt{CV} per dataset. For summarization, we calculate \texttt{CV} of the correlations over the 4 stopword settings for each criterion 
on each dataset.\footnote{TAC datasets have human judgements according to two criteria: Responsivenss and Pyramid; details are given in Section \ref{sec:dataset_description}.}
\paragraph{Results}
\label{re:stopwords}

\begin{figure}[t]
    \centering
    \includegraphics[width=\columnwidth]{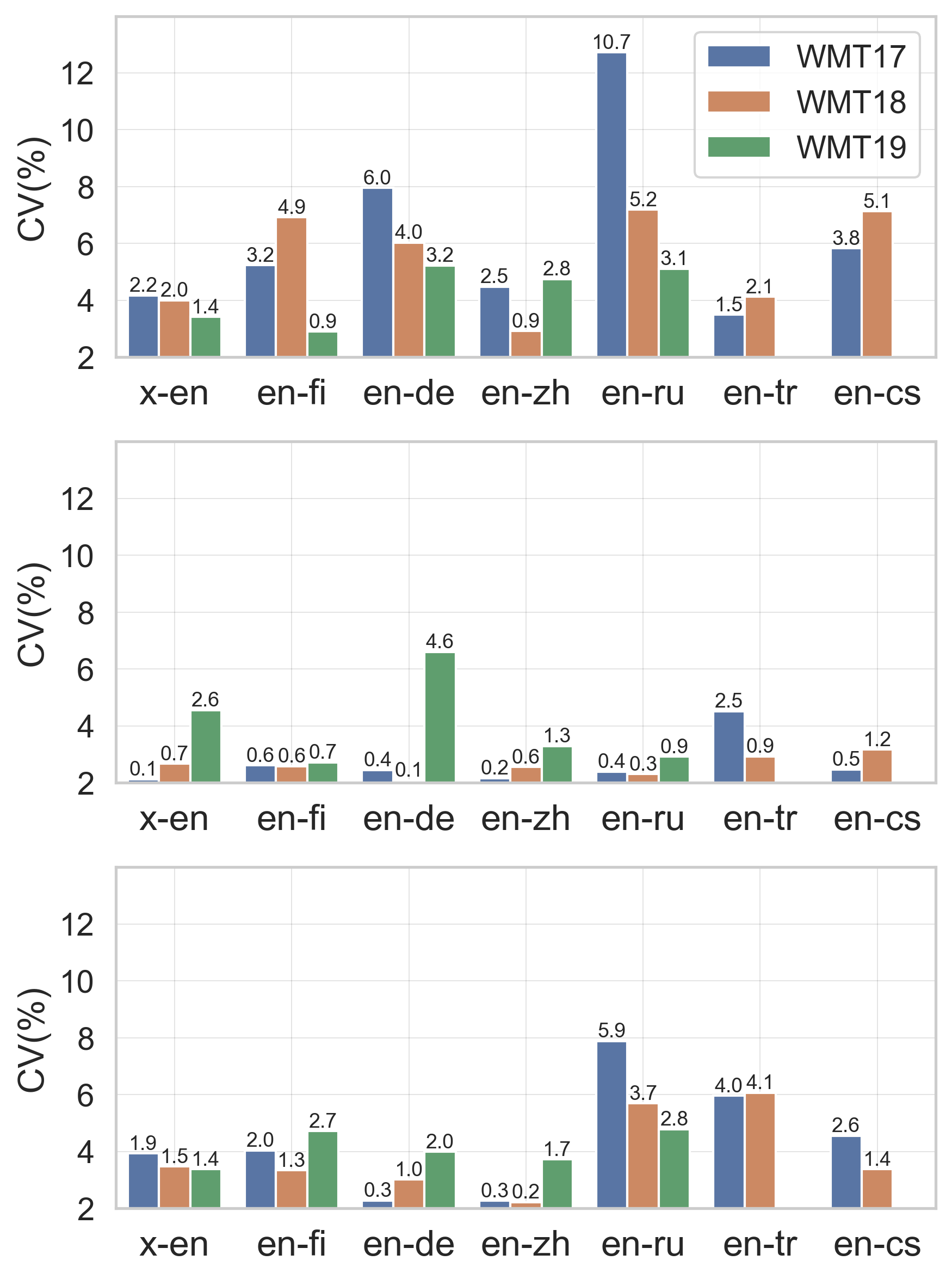}
    \caption{From top to bottom: \texttt{CV}$_{\text{STOP}}$, \texttt{CV}$_{\text{IDF}}$, \texttt{CV}$_{\text{SUB}}$. WMT17-19, segment-level evaluation, MoverScore-1. x-en denotes the average results on all to-English language pairs (where metrics operate on English texts).}
    \label{fig:stopwords}
\end{figure}

On segment-level MT, as Figure \ref{fig:stopwords} (top) shows,  
the sensitivity  
\sen{varies across} datasets and languages. 
Most of the \texttt{CV}$_{\text{STOP}}$ are in range of 2-4\%. This leads to 
6-11\%
absolute variation of the metric performance when the average correlation is, for example, 0.7 (95\% confidence interval). 
For some datasets and languages, the variation is even more pronounced: for example, for Russian on WMT17, the \texttt{CV}$_{\text{STOP}}$ is above 10\%. 

\yc{
Among the examined metrics, 
MoverScore-2 behaves slightly more sensitively than MoverScore-1, whereas BERTScore-F1 is much more sensitive than MoverScore-1 on Chinese and English.  
Compared to other tasks, 
stopwords removal has the largest (but negative) 
impact in segment-level MT evaluation (cf.\ Section \ref{sec:stop_best}).
}

\begin{figure}[t]
    \centering
    \subfigure[RD(dis,ori)]{
    \includegraphics[width=\columnwidth]{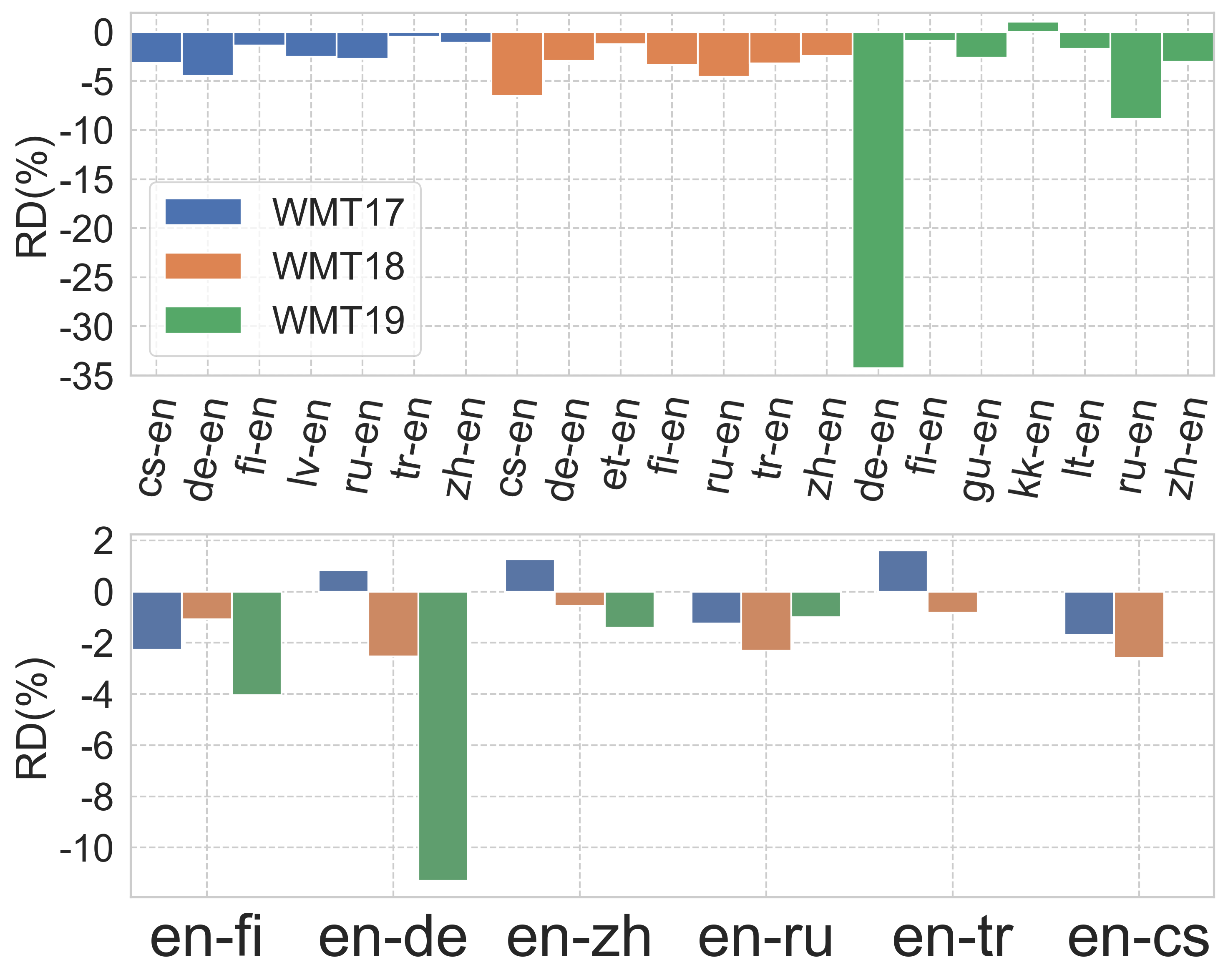}
    }
    \subfigure[RD(dis,pr)]{
    \includegraphics[width=\columnwidth]{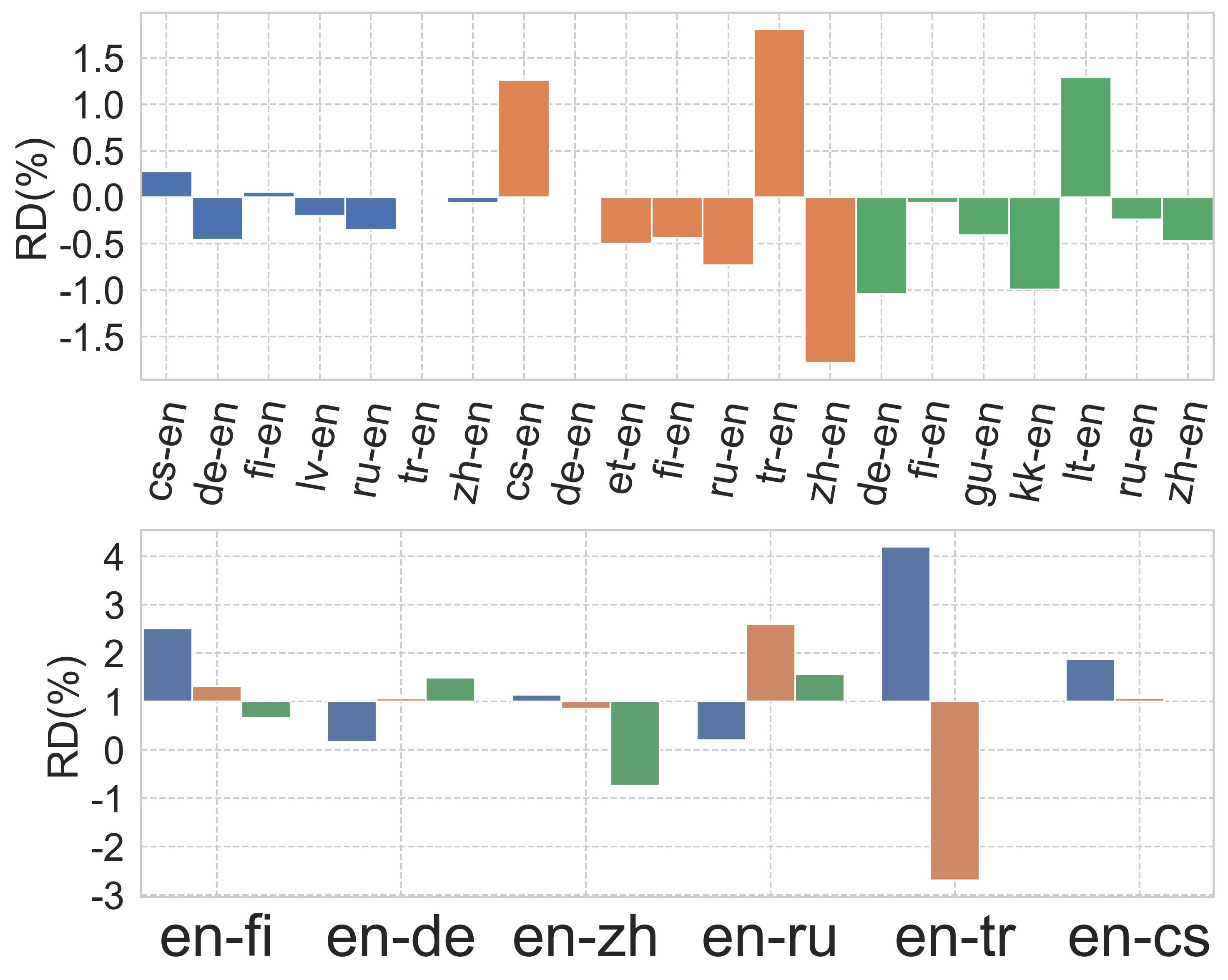}
    }
    \caption{\texttt{RD}(dis,ori) \sen{(for IDF weighting)}, \texttt{RD}(dis,pr) \sen{(for punctuation removal)}. WMT17-19, segment-level evaluation, MoverScore-1. The top graphs in (a) and (b) are the results on to-English language pairs (where metrics operate on English texts), whereas the bottom ones are those on from-English language pairs (where metrics operate on texts in other languages).}
    \label{fig:mover1_rd_seg}
\end{figure}

\subsection{IDF-weighting}
\label{idf set}
In this test, we first disable IDF-weighting for the evaluation metrics (idf$_{\text{dis}}$), and compare the metric performance to that when applying original IDF-weighting\footnote{The original IDF weights for MoverScore are extracted from the reference and hypothesis corpus; those for BERTScore are computed using the reference corpus.
} (idf$_{\text{ori}}$) by calculating the \texttt{RD} between them. We denote this statistic as \texttt{RD}(dis,ori); negative values indicate idf$_{\text{ori}}$ works better and vice versa. 
Next, to inspect the sensitivity to varying IDF-weighting corpora, 
we apply IDF-weighting from four randomly generated corpora to the evaluation metrics additionally (idf$_{\text{rand}}$):  
each corpus consists of 2k English segments randomly selected from the concatenated corpus of all tested datasets. 
The corresponding variability of the metric performance is quantified by the \texttt{CV} of the correlations with humans over the 5 IDF-weighting corpus selections (idf$_{\text{ori}}$ + 4 idf$_{\text{rand}}$), marked with \texttt{CV}$_{\text{IDF}}$. 
We examine the sensitivity regarding IDF-weighting of MoverScore-1, MoverScore-2, and BERTScore-F1. 
Subsequently, we test the IDF-weighting from  
large-scale corpora (idf$_{\text{large}}$). These corpora are obtained from Hugging Face Datasets.\footnote{\url{https://huggingface.co/datasets}. The corpora used here are listed in Section \ref{sec:idf_corpora}.}

\paragraph{Results}\label{re:idf}

\final{As seen in Figure \ref{fig:mover1_rd_seg}(a)}, 
\texttt{RD}(dis,ori) is positive on only one to-English language pair (WMT19 kk-en), 
but on three from-English language pairs (WMT17 en-de, en-zh, and en-tr). Overall, IDF-weighting is thus beneficial. 
The maximal performance drops  
are on WMT19 de-en ( 
>35\%) and 
en-de 
( 
>10\%), respectively. 
Most \texttt{RD}(dis,ori) have absolute values 
<5\%. This means, suppose the correlation is 0.7, the performance can fall by around 0.035 because of disabling IDF-weighting. 

Next, \texttt{CV}$_{\text{IDF}}$ for segment-level MT is presented in Figure \ref{fig:stopwords} (middle). 
In English evaluation, the maximal variation is also caused by the result for de-en in WMT19, where idf$_{\text{ori}}$ yields considerably better result than idf$_{\text{rand}}$ (0.22 vs. 0.17 Kendall's $\tau$).
While en-de has \texttt{CV} values above 4.5\%, 
most \texttt{CV}$_{\text{IDF}}$ are smaller than 1\%. 

\yc{
BERTScore-F1 is less sensitive to IDF-weighting than both MoverScore variants. Among the evaluation tasks, the metrics are again most sensitive on segment-level MT, where for English, 
idf$_{\text{ori}}$ works best for MoverScore (even idf$_{\text{large}}$ cannot 
improve its performance), 
while 
idf$_{\text{rand}}$ and idf$_{\text{ori}}$ are almost equally effective for BERTScore-F1 (cf. Section \ref{sec:idf}).}

\subsection{Subwords \& Punctuation}

In this experiment, we evaluate the sensitivity 
to 
\sen{(i)} subword selection and \sen{(ii)} punctuation removal (PR).  (i) In addition to the original two selections of subwords (keeping the first subword and keeping all subwords),  
we also  
average the embeddings of the subwords in a word to get the word-level BERT representations. 
To quantify the sensitivity to subword selection, we calculate \texttt{CV} of the correlations with humans over the 3 subword selections, denoted as \texttt{CV}$_{\text{SUB}}$. 
(ii) We measure the performance change from using to disabling PR by calculating the \texttt{RD} between them, which we denote as 
\texttt{RD}(dis,pr); negative values indicate MoverScore with PR performs better and vice versa. 
We inspect the corresponding sensitivity of MoverScore-1. 

\paragraph{Results}
\label{re:sub}

\final{Figure \ref{fig:mover1_rd_seg}(b)}  
shows that 
most \texttt{RD}(dis,pr) have absolute values  
<1\%, while both values for en-tr are  
>3\%.
Further, the  
\texttt{CV}$_{\text{SUB}}$ for segment-level MT is presented
in Figure \ref{fig:stopwords} (bottom). 
The average \texttt{CV}$_{\text{SUB}}$ over all datasets for most languages are  
<2\%, whereas 
highly inflectional languages such as Turkish and Russian are considerably more sensitive, with average values  
>4\%. 

\yc{ 
Similar as  
for stopwords and IDF weighting,  
MoverScore-1 behaves most sensitively on segment-level MT, where the default configuration of PR and subwords, which uses the first subword and removes punctuations, works  
best for English. However, for other languages, only in 2 out of 16 cases is it best to select the default configuration (cf. Section \ref{sec:sub_best}). As the authors of MoverScore only reported the results on English data, they may thus select  
an optimal preprocessing strategy only for that case. 
}

\subsection{Discussion}
\label{sec:conclusion}
We summarize the findings from the previous experiments along 4 dimensions.

\textbf{Evaluation Tasks}: 
    Among the considered NLG tasks,  
    BERT-based evaluation metrics are more likely to generate inconsistent scores in segment-level MT evaluation. Their sensitivity 
    is less pronounced in system-level MT and summarization. 
    In the latter two cases, average scores are considered, over the translations within one system or over the multiple references.  
    Thus, some of the variation in metric scores  
    will cancel out, leading to a less fluctuating metric performance from varying preprocessing schemes. 
 \textbf{Evaluation metrics}:
    Among the two variants of MoverScore, 
    MoverScore-2 are more 
    sensitive to parameter settings. 
    BERTScore-F1 behaves less sensitively to IDF-weighting than MoverScore while it behaves much more sensitively to stopwords in the evaluation of Chinese and English compared with MoverScore-1. 
\textbf{Languages}: Overall, the considered evaluation metrics 
have different sensitivities 
in different languages. 
Furthermore,  
highly inflectional languages such as Turkish and Russian as well as German often become ``outliers'' or obtain extrema in our experiments. 
\textbf{Importance of factors}:
    Stopwords removal has the largest but mostly negative impact. 
    IDF-weighting positively impacts 
    evaluation metrics in English evaluation but its contribution is much less stable in the evaluation of other languages. 
    MoverScore benefits from subwords and punctuation removal in segment-level MT evaluation for English, but on other tasks or for other languages, no configuration of PR and subword selection consistently performs best.

\section{Conclusion}
\vspace{-0.15cm}
We investigated reproducibility for BERT-based evaluation metrics, finding several problematic aspects, including using heavy undocumented preprocessing, reporting lower scores for competitors, selective evaluation on 
datasets, and copying correlation scores from wrong indices.  
Our findings cast some doubts on previously reported results and findings, i.e., whether more the complex alignment schemes are really more effective than the greedy alignment of BERTScore.
\se{In terms of preprocessing, we found that it can have a large effect depending (a.o.) on the languages and tasks involved.}
\final{For a fairer comparison between metrics, we recommend to (1) additionally report the results on the datasets that the competitors used, (2) check whether the used versions of the competitor metrics can obtain comparable results as in the original papers, and (3) minimize the role of preprocessing (ideally employing uniform preprocessing across metrics).}
On the positive side, as authors are nowadays much more willing to publish their resources, it is considerably easier to spot such problems, which may also be one reason why critique papers such as ours have become more popular in the last few years \citep{Beese2022DetectingSI}. 
In a wider context, our paper contributes to addressing the ``cracked foundations'' of evaluation for text generation \citep{gehrmann2022repairing} and to better understanding their limitations \citep{Leiter2022TowardsEE}.  

In the future, we would like to reproduce more recent BERT-based metrics --- e.g., with other aggregation mechanisms \citep{chen-etal-2020-improving-text},  normalization schemes \citep{zhao-etal-2021-inducing}, different design choices \citep{yuan2021bartscore, chen2022menli}, \final{or metrics that use supervision \citep{rei2020comet, sellam-etal-2020-bleurt, Rony2022RoMeAR}} --- to obtain a broader assessment of reproducibility issues in this context. We would also like to quantify, at a larger scale, the bias in research induced from overestimating one's own model vis-\`a-vis competitor models.

\section{Limitations}
Limitations of our work include (1) a limited number of explored evaluation metrics, (2) 
a restricted focus on MT only 
 and (3) reliance on author-provided reproduction resources.

(1) 
Although we did point out very important issues, 
we only reproduced four metrics. Further, the sensitivity analysis only concerned two evaluation metrics. 
In the future, we would like to include more reproducibility studies on recent BERT-based evaluation metrics for a broader analysis. It is possible that our particular sample is representative of more severe underlying problems in the community or that 
it is particularly affected by reproducibility issues. 

(2) 
Our reproduction attempts, with the exception of MoverScore, focused only on MT. For example, the authors of BaryScore also reported results on summarization, IC, and D2T generation, 
which (for computational costs) we did not considered in this work. While we believe that our findings generalize from MT to other tasks, we did not confirm this expectation experimentally. 

(3) Our reproduction attempts were mainly based on the author-provided resources, such as the code and datasets they released, with which we could obtain comparable results in most instances. Nevertheless, we did not investigate their legitimacy, e.g., whether the implementation of the approach is in accordance with the description in its paper or whether the datasets uploaded by the authors are the official ones, etc.

\section{Acknowledgement}
\final{We thank the Ubiquitous Knowledge Processing (UKP) Lab at TU Darmstadt for providing the computing resources. We thank the reviewers for their thoughtful feedback, which improved the final version of the paper. We also thank the authors of MoverScore and BaryScore for their feedback on the initial version of this paper.}

\bibliography{anthology,custom}

\begin{thebibliography}{60}
\expandafter\ifx\csname natexlab\endcsname\relax\def\natexlab#1{#1}\fi

\bibitem[{Agueh and Carlier(2011)}]{agueh2011barycenters}
Martial Agueh and Guillaume Carlier. 2011.
\newblock Barycenters in the wasserstein space.
\newblock \emph{SIAM Journal on Mathematical Analysis}, 43(2):904--924.

\bibitem[{Anderson et~al.(2016)Anderson, Fernando, Johnson, and
  Gould}]{anderson2016spice}
Peter Anderson, Basura Fernando, Mark Johnson, and Stephen Gould. 2016.
\newblock Spice: Semantic propositional image caption evaluation.
\newblock In \emph{European conference on computer vision}, pages 382--398.
  Springer.

\bibitem[{Baker(2016)}]{baker2016reproducibility}
Monya Baker. 2016.
\newblock Reproducibility crisis.
\newblock \emph{Nature}, 533(26):353--66.

\bibitem[{Beese et~al.(2022)Beese, Altunbacs, Guzeler, and
  Eger}]{Beese2022DetectingSI}
Dominik Beese, Begum Altunbacs, Gorkem Guzeler, and Steffen Eger. 2022.
\newblock Detecting stance in scientific papers: Did we get more negative
  recently?
\newblock \emph{ArXiv}, abs/2202.13610.

\bibitem[{Belouadi and Eger(2022)}]{Belouadi2022USCOREAE}
Jonas Belouadi and Steffen Eger. 2022.
\newblock Uscore: An effective approach to fully unsupervised evaluation
  metrics for machine translation.
\newblock \emph{ArXiv}, abs/2202.10062.

\bibitem[{Belz et~al.(2021)Belz, Agarwal, Shimorina, and
  Reiter}]{belz-etal-2021-systematic}
Anya Belz, Shubham Agarwal, Anastasia Shimorina, and Ehud Reiter. 2021.
\newblock \href {https://doi.org/10.18653/v1/2021.eacl-main.29} {A systematic
  review of reproducibility research in natural language processing}.
\newblock In \emph{Proceedings of the 16th Conference of the European Chapter
  of the Association for Computational Linguistics: Main Volume}, pages
  381--393, Online. Association for Computational Linguistics.

\bibitem[{Belz et~al.(2022)Belz, Popovi{\'c}, and Mille}]{belz2022quantified}
Anya Belz, Maja Popovi{\'c}, and Simon Mille. 2022.
\newblock Quantified reproducibility assessment of nlp results.
\newblock \emph{arXiv preprint arXiv:2204.05961}.

\bibitem[{Bird et~al.(2009)Bird, Klein, and Loper}]{bird2009natural}
Steven Bird, Ewan Klein, and Edward Loper. 2009.
\newblock \emph{Natural language processing with Python: analyzing text with
  the natural language toolkit}.
\newblock " O'Reilly Media, Inc.".

\bibitem[{Bojar et~al.(2016)Bojar, Graham, Kamran, and Stanojevi{\'c}}]{wmt16}
Ond{\v{r}}ej Bojar, Yvette Graham, Amir Kamran, and Milo{\v{s}} Stanojevi{\'c}.
  2016.
\newblock Results of the wmt16 metrics shared task.
\newblock In \emph{Proceedings of the First Conference on Machine Translation:
  Volume 2, Shared Task Papers}, pages 199--231.

\bibitem[{Bojar et~al.(2017)Bojar, Graham, and Kamran}]{wmt17}
Ondřej Bojar, Yvette Graham, and Amir Kamran. 2017.
\newblock \href {https://doi.org/10.18653/v1/W17-4755} {Results of the wmt17
  metrics shared task}.
\newblock pages 489--513.

\bibitem[{Chen et~al.(2020)Chen, Ding, Levinboim, and
  Soricut}]{chen-etal-2020-improving-text}
Xi~Chen, Nan Ding, Tomer Levinboim, and Radu Soricut. 2020.
\newblock \href {https://doi.org/10.18653/v1/2020.eval4nlp-1.6} {Improving text
  generation evaluation with batch centering and tempered word mover distance}.
\newblock In \emph{Proceedings of the First Workshop on Evaluation and
  Comparison of NLP Systems}, pages 51--59, Online. Association for
  Computational Linguistics.

\bibitem[{Chen and Eger(2022)}]{chen2022menli}
Yanran Chen and Steffen Eger. 2022.
\newblock Menli: Robust evaluation metrics from natural language inference.
\newblock \emph{arXiv preprint arXiv:2208.07316}.

\bibitem[{Cohen et~al.(2018)Cohen, Xia, Zweigenbaum, Callahan, Hargraves, Goss,
  Ide, N{\'e}v{\'e}ol, Grouin, and Hunter}]{Cohen2018ThreeDO}
Kevin~Bretonnel Cohen, Jingbo Xia, Pierre Zweigenbaum, Tiffany~J. Callahan,
  Orin Hargraves, Foster~R. Goss, Nancy Ide, Aur{\'e}lie N{\'e}v{\'e}ol, Cyril
  Grouin, and Lawrence~E. Hunter. 2018.
\newblock Three dimensions of reproducibility in natural language processing.
\newblock \emph{LREC ... International Conference on Language Resources \&
  Evaluation : [proceedings]. International Conference on Language Resources
  and Evaluation}, 2018:156--165.

\bibitem[{Colombo et~al.(2021)Colombo, Staerman, Clavel, and
  Piantanida}]{colombo2021automatic}
Pierre Colombo, Guillaume Staerman, Chloe Clavel, and Pablo Piantanida. 2021.
\newblock Automatic text evaluation through the lens of wasserstein
  barycenters.
\newblock \emph{arXiv preprint arXiv:2108.12463}.

\bibitem[{Cui et~al.(2018)Cui, Yang, Veit, Huang, and
  Belongie}]{cui2018learning}
Yin Cui, Guandao Yang, Andreas Veit, Xun Huang, and Serge Belongie. 2018.
\newblock Learning to evaluate image captioning.
\newblock In \emph{Proceedings of the IEEE conference on computer vision and
  pattern recognition}, pages 5804--5812.

\bibitem[{Cuturi(2013)}]{cuturi2013sinkhorn}
Marco Cuturi. 2013.
\newblock Sinkhorn distances: Lightspeed computation of optimal transport.
\newblock \emph{Advances in neural information processing systems}, 26.

\bibitem[{Dakota and K{\"u}bler(2017)}]{Dakota2017TowardsRI}
Daniel Dakota and Sandra K{\"u}bler. 2017.
\newblock Towards replicability in parsing.
\newblock In \emph{RANLP}.

\bibitem[{Devlin et~al.(2018)Devlin, Chang, Lee, and
  Toutanova}]{devlin2018bert}
Jacob Devlin, Ming-Wei Chang, Kenton Lee, and Kristina Toutanova. 2018.
\newblock Bert: Pre-training of deep bidirectional transformers for language
  understanding.
\newblock \emph{arXiv preprint arXiv:1810.04805}.

\bibitem[{Fokkens et~al.(2013)Fokkens, van Erp, Postma, Pedersen, Vossen, and
  Freire}]{Fokkens2013OffspringFR}
Antske Fokkens, Marieke van Erp, Marten Postma, Ted Pedersen, P.~Vossen, and
  Nuno Freire. 2013.
\newblock Offspring from reproduction problems: What replication failure
  teaches us.
\newblock In \emph{ACL}.

\bibitem[{Fomicheva et~al.(2020{\natexlab{a}})Fomicheva, Sun, Yankovskaya,
  Blain, Guzm{\'a}n, Fishel, Aletras, Chaudhary, and
  Specia}]{Fomicheva2020UnsupervisedQE}
M.~Fomicheva, Shuo Sun, Lisa Yankovskaya, F.~Blain, Francisco Guzm{\'a}n,
  M.~Fishel, Nikolaos Aletras, Vishrav Chaudhary, and Lucia Specia.
  2020{\natexlab{a}}.
\newblock Unsupervised quality estimation for neural machine translation.
\newblock \emph{Transactions of the Association for Computational Linguistics},
  8:539--555.

\bibitem[{Fomicheva et~al.(2020{\natexlab{b}})Fomicheva, Sun, Fonseca, Zerva,
  Blain, Chaudhary, Guzm{\'a}n, Lopatina, Specia, and
  Martins}]{fomicheva2020mlqe}
Marina Fomicheva, Shuo Sun, Erick Fonseca, Chrysoula Zerva, Fr{\'e}d{\'e}ric
  Blain, Vishrav Chaudhary, Francisco Guzm{\'a}n, Nina Lopatina, Lucia Specia,
  and Andr{\'e}~FT Martins. 2020{\natexlab{b}}.
\newblock Mlqe-pe: A multilingual quality estimation and post-editing dataset.
\newblock \emph{arXiv preprint arXiv:2010.04480}.

\bibitem[{Foundation()}]{wikidump}
Wikimedia Foundation.
\newblock \href {https://dumps.wikimedia.org} {Wikimedia downloads}.

\bibitem[{Gao et~al.(2021)Gao, Eger, Zhao, Lertvittayakumjorn, and
  Fomicheva}]{eval4nlp-2021-evaluation}
Yang Gao, Steffen Eger, Wei Zhao, Piyawat Lertvittayakumjorn, and Marina
  Fomicheva, editors. 2021.
\newblock \href {https://aclanthology.org/2021.eval4nlp-1.0} {\emph{Proceedings
  of the 2nd Workshop on Evaluation and Comparison of NLP Systems}}.
  Association for Computational Linguistics, Punta Cana, Dominican Republic.

\bibitem[{Gehrmann et~al.(2022)Gehrmann, Clark, and
  Sellam}]{gehrmann2022repairing}
Sebastian Gehrmann, Elizabeth Clark, and Thibault Sellam. 2022.
\newblock \href {http://arxiv.org/abs/2202.06935} {Repairing the cracked
  foundation: A survey of obstacles in evaluation practices for generated
  text}.

\bibitem[{Guo et~al.(2020)Guo, Dai, Vrande{\v{c}}i{\'c}, and
  Al-Rfou}]{guo-etal-2020-wiki}
Mandy Guo, Zihang Dai, Denny Vrande{\v{c}}i{\'c}, and Rami Al-Rfou. 2020.
\newblock \href {https://aclanthology.org/2020.lrec-1.297} {{W}iki-40{B}:
  Multilingual language model dataset}.
\newblock In \emph{Proceedings of the 12th Language Resources and Evaluation
  Conference}, pages 2440--2452, Marseille, France. European Language Resources
  Association.

\bibitem[{Guo et~al.(2018)Guo, Ruan, and Hu}]{guo-etal-2018-meteor}
Yinuo Guo, Chong Ruan, and Junfeng Hu. 2018.
\newblock \href {https://doi.org/10.18653/v1/W18-6454} {{M}eteor++:
  Incorporating copy knowledge into machine translation evaluation}.
\newblock In \emph{Proceedings of the Third Conference on Machine Translation:
  Shared Task Papers}, pages 740--745, Belgium, Brussels. Association for
  Computational Linguistics.

\bibitem[{Honnibal and Montani(2017)}]{spacy2}
Matthew Honnibal and Ines Montani. 2017.
\newblock {spaCy 2}: Natural language understanding with {B}loom embeddings,
  convolutional neural networks and incremental parsing.
\newblock To appear.

\bibitem[{Kusner et~al.(2015)Kusner, Sun, Kolkin, and
  Weinberger}]{kusner2015word}
Matt Kusner, Yu~Sun, Nicholas Kolkin, and Kilian Weinberger. 2015.
\newblock From word embeddings to document distances.
\newblock In \emph{International conference on machine learning}, pages
  957--966. PMLR.

\bibitem[{Leiter et~al.(2022)Leiter, Lertvittayakumjorn, Fomicheva, Zhao, Gao,
  and Eger}]{Leiter2022TowardsEE}
Christoph Leiter, Piyawat Lertvittayakumjorn, M.~Fomicheva, Wei Zhao, Yang Gao,
  and Steffen Eger. 2022.
\newblock Towards explainable evaluation metrics for natural language
  generation.

\bibitem[{Lo(2019)}]{lo-2019-yisi}
Chi-kiu Lo. 2019.
\newblock \href {https://doi.org/10.18653/v1/W19-5358} {{Y}i{S}i - a unified
  semantic {MT} quality evaluation and estimation metric for languages with
  different levels of available resources}.
\newblock In \emph{Proceedings of the Fourth Conference on Machine Translation
  (Volume 2: Shared Task Papers, Day 1)}, pages 507--513, Florence, Italy.
  Association for Computational Linguistics.

\bibitem[{Ma et~al.(2018)Ma, Bojar, and Graham}]{wmt18}
Qingsong Ma, Ondřej Bojar, and Yvette Graham. 2018.
\newblock \href {https://doi.org/10.18653/v1/W18-6450} {Results of the wmt18
  metrics shared task: Both characters and embeddings achieve good
  performance}.
\newblock pages 671--688.

\bibitem[{Ma et~al.(2019)Ma, Wei, Bojar, and Graham}]{wmt19}
Qingsong Ma, Johnny Wei, Ond{\v{r}}ej Bojar, and Yvette Graham. 2019.
\newblock Results of the wmt19 metrics shared task: Segment-level and strong mt
  systems pose big challenges.
\newblock In \emph{Proceedings of the Fourth Conference on Machine Translation
  (Volume 2: Shared Task Papers, Day 1)}, pages 62--90.

\bibitem[{Maas et~al.(2011)Maas, Daly, Pham, Huang, Ng, and
  Potts}]{maas-EtAl:2011:ACL-HLT2011}
Andrew~L. Maas, Raymond~E. Daly, Peter~T. Pham, Dan Huang, Andrew~Y. Ng, and
  Christopher Potts. 2011.
\newblock \href {http://www.aclweb.org/anthology/P11-1015} {Learning word
  vectors for sentiment analysis}.
\newblock In \emph{Proceedings of the 49th Annual Meeting of the Association
  for Computational Linguistics: Human Language Technologies}, pages 142--150,
  Portland, Oregon, USA. Association for Computational Linguistics.

\bibitem[{Mach{\'a}{\v{c}}ek and Bojar(2014)}]{machacek-bojar-2014-results}
Matou{\v{s}} Mach{\'a}{\v{c}}ek and Ond{\v{r}}ej Bojar. 2014.
\newblock \href {https://doi.org/10.3115/v1/W14-3336} {Results of the {WMT}14
  metrics shared task}.
\newblock In \emph{Proceedings of the Ninth Workshop on Statistical Machine
  Translation}, pages 293--301, Baltimore, Maryland, USA. Association for
  Computational Linguistics.

\bibitem[{Mairesse et~al.(2010)Mairesse, Gasic, Jurcicek, Keizer, Thomson, Yu,
  and Young}]{mairesse2010phrase}
Fran{\c{c}}ois Mairesse, Milica Gasic, Filip Jurcicek, Simon Keizer, Blaise
  Thomson, Kai Yu, and Steve Young. 2010.
\newblock Phrase-based statistical language generation using graphical models
  and active learning.
\newblock In \emph{Proceedings of the 48th Annual Meeting of the Association
  for Computational Linguistics}, pages 1552--1561.

\bibitem[{Marie et~al.(2021)Marie, Fujita, and
  Rubino}]{marie-etal-2021-scientific}
Benjamin Marie, Atsushi Fujita, and Raphael Rubino. 2021.
\newblock \href {https://doi.org/10.18653/v1/2021.acl-long.566} {Scientific
  credibility of machine translation research: A meta-evaluation of 769
  papers}.
\newblock In \emph{Proceedings of the 59th Annual Meeting of the Association
  for Computational Linguistics and the 11th International Joint Conference on
  Natural Language Processing (Volume 1: Long Papers)}, pages 7297--7306,
  Online. Association for Computational Linguistics.

\bibitem[{Mathur et~al.(2020)Mathur, Baldwin, and
  Cohn}]{mathur-etal-2020-tangled}
Nitika Mathur, Timothy Baldwin, and Trevor Cohn. 2020.
\newblock \href {https://doi.org/10.18653/v1/2020.acl-main.448} {Tangled up in
  {BLEU}: Reevaluating the evaluation of automatic machine translation
  evaluation metrics}.
\newblock In \emph{Proceedings of the 58th Annual Meeting of the Association
  for Computational Linguistics}, pages 4984--4997, Online. Association for
  Computational Linguistics.

\bibitem[{Merity et~al.(2016)Merity, Xiong, Bradbury, and
  Socher}]{merity2016pointer}
Stephen Merity, Caiming Xiong, James Bradbury, and Richard Socher. 2016.
\newblock \href {http://arxiv.org/abs/1609.07843} {Pointer sentinel mixture
  models}.

\bibitem[{Mieskes(2017)}]{Mieskes2017AQS}
Margot Mieskes. 2017.
\newblock A quantitative study of data in the nlp community.
\newblock In \emph{EthNLP@EACL}.

\bibitem[{Novikova et~al.(2017)Novikova, Du{\v{s}}ek, Cercas~Curry, and
  Rieser}]{novikova-etal-2017-need}
Jekaterina Novikova, Ond{\v{r}}ej Du{\v{s}}ek, Amanda Cercas~Curry, and Verena
  Rieser. 2017.
\newblock \href {https://doi.org/10.18653/v1/D17-1238} {Why we need new
  evaluation metrics for {NLG}}.
\newblock In \emph{Proceedings of the 2017 Conference on Empirical Methods in
  Natural Language Processing}, pages 2241--2252, Copenhagen, Denmark.
  Association for Computational Linguistics.

\bibitem[{Papineni et~al.(2002)Papineni, Roukos, Ward, and
  Zhu}]{Papineni2002BleuAM}
Kishore Papineni, Salim Roukos, Todd Ward, and Wei-Jing Zhu. 2002.
\newblock Bleu: a method for automatic evaluation of machine translation.
\newblock In \emph{ACL}.

\bibitem[{Post(2018)}]{post2018call}
Matt Post. 2018.
\newblock \href {https://doi.org/10.18653/v1/W18-6319} {A call for clarity in
  reporting {BLEU} scores}.
\newblock In \emph{Proceedings of the Third Conference on Machine Translation:
  Research Papers}, pages 186--191, Brussels, Belgium. Association for
  Computational Linguistics.

\bibitem[{Rei et~al.(2020)Rei, Stewart, Farinha, and Lavie}]{rei2020comet}
Ricardo Rei, Craig Stewart, Ana~C Farinha, and Alon Lavie. 2020.
\newblock \href {http://arxiv.org/abs/2009.09025} {Comet: A neural framework
  for mt evaluation}.

\bibitem[{Reimers and Gurevych(2020)}]{reimers-gurevych-2020-making}
Nils Reimers and Iryna Gurevych. 2020.
\newblock \href {https://doi.org/10.18653/v1/2020.emnlp-main.365} {Making
  monolingual sentence embeddings multilingual using knowledge distillation}.
\newblock In \emph{Proceedings of the 2020 Conference on Empirical Methods in
  Natural Language Processing (EMNLP)}, pages 4512--4525, Online. Association
  for Computational Linguistics.

\bibitem[{Rony et~al.(2022)Rony, Kovriguina, Chaudhuri, Usbeck, and
  Lehmann}]{Rony2022RoMeAR}
Md~Rashad Al~Hasan Rony, Liubov Kovriguina, Debanjan Chaudhuri, Ricardo Usbeck,
  and Jens Lehmann. 2022.
\newblock \href {https://doi.org/10.18653/v1/2022.acl-long.387} {{R}o{M}e: A
  robust metric for evaluating natural language generation}.
\newblock In \emph{Proceedings of the 60th Annual Meeting of the Association
  for Computational Linguistics (Volume 1: Long Papers)}, pages 5645--5657,
  Dublin, Ireland. Association for Computational Linguistics.

\bibitem[{Rubner et~al.(2000)Rubner, Tomasi, and Guibas}]{rubner2000earth}
Yossi Rubner, Carlo Tomasi, and Leonidas~J Guibas. 2000.
\newblock The earth mover's distance as a metric for image retrieval.
\newblock \emph{International journal of computer vision}, 40(2):99--121.

\bibitem[{R{\"u}ckl{\'e} et~al.(2018)R{\"u}ckl{\'e}, Eger, Peyrard, and
  Gurevych}]{rueckle-etal-2018-pmeans}
Andreas R{\"u}ckl{\'e}, Steffen Eger, Maxime Peyrard, and Iryna Gurevych. 2018.
\newblock \href {https://arxiv.org/abs/1803.01400} {Concatenated power mean
  word embeddings as universal cross-lingual sentence representations}.
\newblock \emph{arXiv}.

\bibitem[{Sellam et~al.(2020)Sellam, Das, and Parikh}]{sellam-etal-2020-bleurt}
Thibault Sellam, Dipanjan Das, and Ankur Parikh. 2020.
\newblock \href {https://doi.org/10.18653/v1/2020.acl-main.704} {{BLEURT}:
  Learning robust metrics for text generation}.
\newblock In \emph{Proceedings of the 58th Annual Meeting of the Association
  for Computational Linguistics}, pages 7881--7892, Online. Association for
  Computational Linguistics.

\bibitem[{Song et~al.(2021)Song, Zhao, and Specia}]{song-etal-2021-sentsim}
Yurun Song, Junchen Zhao, and Lucia Specia. 2021.
\newblock \href {https://doi.org/10.18653/v1/2021.naacl-main.252} {{S}ent{S}im:
  Crosslingual semantic evaluation of machine translation}.
\newblock In \emph{Proceedings of the 2021 Conference of the North American
  Chapter of the Association for Computational Linguistics: Human Language
  Technologies}, pages 3143--3156, Online. Association for Computational
  Linguistics.

\bibitem[{Specia et~al.(2020)Specia, Blain, Fomicheva, Fonseca, Chaudhary,
  Guzm{\'a}n, and Martins}]{Specia2020FindingsOT}
Lucia Specia, F.~Blain, M.~Fomicheva, E.~Fonseca, Vishrav Chaudhary, Francisco
  Guzm{\'a}n, and Andr{\'e} F.~T. Martins. 2020.
\newblock Findings of the wmt 2020 shared task on quality estimation.
\newblock In \emph{WMT@EMNLP}.

\bibitem[{Stanojevi{\'c} et~al.(2015)Stanojevi{\'c}, Kamran, Koehn, and
  Bojar}]{wmt15}
Milo{\v{s}} Stanojevi{\'c}, Amir Kamran, Philipp Koehn, and Ond{\v{r}}ej Bojar.
  2015.
\newblock Results of the wmt15 metrics shared task.
\newblock In \emph{Proceedings of the Tenth Workshop on Statistical Machine
  Translation}, pages 256--273.

\bibitem[{Thoma(2018)}]{thoma_martin_2018_841984}
Martin Thoma. 2018.
\newblock \href {https://doi.org/10.5281/zenodo.841984} {{WiLI-2018 - Wikipedia
  Language Identification database}}.

\bibitem[{Wang et~al.(2018)Wang, Singh, Michael, Hill, Levy, and
  Bowman}]{wang-etal-2018-glue}
Alex Wang, Amanpreet Singh, Julian Michael, Felix Hill, Omer Levy, and Samuel
  Bowman. 2018.
\newblock \href {https://doi.org/10.18653/v1/W18-5446} {{GLUE}: A multi-task
  benchmark and analysis platform for natural language understanding}.
\newblock In \emph{Proceedings of the 2018 {EMNLP} Workshop {B}lackbox{NLP}:
  Analyzing and Interpreting Neural Networks for {NLP}}, pages 353--355,
  Brussels, Belgium. Association for Computational Linguistics.

\bibitem[{Wen et~al.(2015)Wen, Gasic, Mrksic, Su, Vandyke, and
  Young}]{wen2015semantically}
Tsung-Hsien Wen, Milica Gasic, Nikola Mrksic, Pei-Hao Su, David Vandyke, and
  Steve Young. 2015.
\newblock \href {http://arxiv.org/abs/1508.01745} {Semantically conditioned
  lstm-based natural language generation for spoken dialogue systems}.

\bibitem[{Wieling et~al.(2018)Wieling, Rawee, and van
  Noord}]{wieling-etal-2018-squib}
Martijn Wieling, Josine Rawee, and Gertjan van Noord. 2018.
\newblock \href {https://doi.org/10.1162/coli_a_00330} {{S}quib:
  Reproducibility in computational linguistics: Are we willing to share?}
\newblock \emph{Computational Linguistics}, 44(4):641--649.

\bibitem[{Yuan et~al.(2021)Yuan, Neubig, and Liu}]{yuan2021bartscore}
Weizhe Yuan, Graham Neubig, and Pengfei Liu. 2021.
\newblock Bartscore: Evaluating generated text as text generation.
\newblock \emph{Advances in Neural Information Processing Systems}, 34.

\bibitem[{Zhang et~al.(2019)Zhang, Kishore, Wu, Weinberger, and
  Artzi}]{zhang2019bertscore}
Tianyi Zhang, Varsha Kishore, Felix Wu, Kilian~Q Weinberger, and Yoav Artzi.
  2019.
\newblock Bertscore: Evaluating text generation with bert.
\newblock \emph{arXiv preprint arXiv:1904.09675}.

\bibitem[{Zhao et~al.(2021)Zhao, Eger, Bjerva, and
  Augenstein}]{zhao-etal-2021-inducing}
Wei Zhao, Steffen Eger, Johannes Bjerva, and Isabelle Augenstein. 2021.
\newblock \href {https://doi.org/10.18653/v1/2021.starsem-1.22} {Inducing
  language-agnostic multilingual representations}.
\newblock In \emph{Proceedings of *SEM 2021: The Tenth Joint Conference on
  Lexical and Computational Semantics}, pages 229--240, Online. Association for
  Computational Linguistics.

\bibitem[{Zhao et~al.(2020)Zhao, Glava{\v{s}}, Peyrard, Gao, West, and
  Eger}]{zhao2020limitations}
Wei Zhao, Goran Glava{\v{s}}, Maxime Peyrard, Yang Gao, Robert West, and
  Steffen Eger. 2020.
\newblock On the limitations of cross-lingual encoders as exposed by
  reference-free machine translation evaluation.
\newblock \emph{arXiv preprint arXiv:2005.01196}.

\bibitem[{Zhao et~al.(2019)Zhao, Peyrard, Liu, Gao, Meyer, and
  Eger}]{zhao2019moverscore}
Wei Zhao, Maxime Peyrard, Fei Liu, Yang Gao, Christian~M Meyer, and Steffen
  Eger. 2019.
\newblock Moverscore: Text generation evaluating with contextualized embeddings
  and earth mover distance.
\newblock \emph{arXiv preprint arXiv:1909.02622}.

\end{thebibliography}
\bibliographystyle{acl_natbib}

\clearpage
\appendix

\section{Appendix}
\label{sec:appendix}

\subsection{Datasets}
\label{sec:dataset_description}
\subsubsection{Machine Translation}

Each WMT dataset contains evaluation data for different pairs of source and translation languages.  
Two types of human judgments serve as the golden standard. The first one is Direct Assessment (DA), which contains human scores for each translation. The second one is DArr, which consists of conclusions about one translation being better than another drawn from DA scores. According to \citet{wmt17}, \citet{wmt18} and \citet{wmt19}, when there is insufficient amount of DA scores for each individual translation (smaller than 15), DArr is then considered. In this work, We follow the official instructions to calculate correlations of evaluation metrics with DA judgments using absolute Pearson's $r$, and with DArr judgments using Kendall's $\tau$-like formulation proposed by \citet{wmt17}. On those datasets, DA always serve as the golden truth for system-level evaluation. For segment-level evaluation, WMT18 and WMT19 use DArr, WMT15 and WMT16 rely on DA, and WMT17 uses DA for all to-English and 2 from-English languages pairs (en-ru and en-zh) and DArr for the remaining from-English language pairs.

\subsubsection{Text Summarization}
Each TAC dataset contains several clusters each with 10 news articles. There are more than 50 system and 4 reference summaries with fewer than 100 words for each article. Each system summary receives 4 human judgements according to two criteria: 1) Pyramid, which reflects the level of content coverage of the summaries; and 2) Responsivenss, which measures the response level to the overall quality of linguistic and content of the summaries. The difference between these two datasets is the fact that TAC2008 contains 48 clusters and summaries from 57 systems, while TAC2009 contains 44 clusters and summaries from 55 systems. \citet{zhao2019moverscore} calculated Pearson and Spearman correlation with summary-level human judgments when evaluating MoverScore. In addition, We compute Kendall correlation as well, allowing for a comparison among the three correlations.

\subsubsection{Image Captioning}
Following \citet{cui2018learning}, \citet{zhao2019moverscore} evaluated MoverScore on the validation set of MSCOCO, which contains roughly 40k images each with 5 reference and 12 system captions. Besides, there are system-level human judgements about 5 criteria: M1-M5 \citep{anderson2016spice}. In the reproduction experiment, following the experiment setup of \citet{zhao2019moverscore}, We calculate Pearson correlation with M1 and M2 scores, which refer to the ratio of captions better or equal to human captions and the ratio of captions indistinguishable from human captions, respectively. 

\subsubsection{Data-to-Text Generation}
There are 202 Meaning Representation (MR) instances in BAGEL and 398 MR instances in SFHOTEL datasets. Multiple references and about two system utterances exist for each MR instance. The datasets provide utterance-level human judgments according to 3 criteria: 1) informativeness, which measures how informative the utterance is; 2) naturalness, which refers to the similarity extent between a system utterance and an native speaker-generated utterance; 3) quality, which reflects the fluency and grammar level of a system utterance \citep{novikova-etal-2017-need}. In the reproduction experiment, We follow \citet{zhao2019moverscore} to calculate Spearman correlation with utterance-level human judgements about these 3 criteria.

\subsection{Reproduction on WMT15-16}\label{sec:repro_wmt15}

Table \ref{tab:repro_wmt1516_spearman} and \ref{tab:repro_wmt1516_kendall} display the reproduced Spearman and Kendall correlations on WMT15 and WMT16. 

\begin{table*}[!htb]
\centering
\resizebox{\textwidth}{!}{%
\begin{tabular}{@{}c|llllll|lllllll@{}}
\toprule
\multicolumn{1}{l|}{} &
  \multicolumn{6}{c|}{WMT15} &
  \multicolumn{7}{c}{WMT16} \\ 
\multicolumn{1}{l|}{} &
  metric &
  \multicolumn{1}{c}{cs-en} &
  \multicolumn{1}{c}{de-en} &
  \multicolumn{1}{c}{fi-en} &
  \multicolumn{1}{c}{ru-en} &
  \multicolumn{1}{c|}{avg} &
  \multicolumn{1}{c}{cs-en} &
  \multicolumn{1}{c}{de-en} &
  \multicolumn{1}{c}{ru-en} &
  \multicolumn{1}{c}{fi-en} &
  \multicolumn{1}{c}{ro-en} &
  \multicolumn{1}{c}{tr-en} &
  \multicolumn{1}{c}{avg} \\ \midrule
 &
  BERT-F1 &
  {\color[HTML]{009901} \textbf{0.750*}} &
  {\color[HTML]{009901} 0.720} &
  {\color[HTML]{009901} 0.731} &
  {\color[HTML]{009901} \textbf{0.712}} &
  {\color[HTML]{009901} 0.728} &
  {\color[HTML]{009901} \textbf{0.749}} &
  {\color[HTML]{FE0000} 0.643} &
  {\color[HTML]{009901} 0.661} &
  {\color[HTML]{009901} \textbf{0.660}} &
  {\color[HTML]{009901} \textbf{0.706}} &
  {\color[HTML]{FE0000} 0.666} &
  {\color[HTML]{009901} \textbf{0.681}} \\
 &
  Mover-1 &
  {\color[HTML]{009901} 0.728} &
  {\color[HTML]{009901} \textbf{0.721}} &
  {\color[HTML]{009901} 0.722} &
  {\color[HTML]{009901} 0.696} &
  {\color[HTML]{009901} 0.717} &
  {\color[HTML]{009901} 0.737} &
  {\color[HTML]{009901} 0.626} &
  {\color[HTML]{009901} \textbf{0.671}} &
  {\color[HTML]{009901} 0.650} &
  {\color[HTML]{009901} 0.696} &
  {\color[HTML]{009901} 0.669} &
  {\color[HTML]{009901} 0.675} \\
 &
  Bary-W &
  {\color[HTML]{009901} 0.747} &
  {\color[HTML]{FE0000} 0.705} &
  {\color[HTML]{009901} \textbf{0.757}} &
  {\color[HTML]{009901} 0.709} &
  {\color[HTML]{009901} \textbf{0.730}} &
  {\color[HTML]{FE0000} 0.730} &
  {\color[HTML]{009901} \textbf{0.650}} &
  {\color[HTML]{FE0000} 0.649} &
  {\color[HTML]{FE0000} 0.660} &
  {\color[HTML]{FE0000} 0.699} &
   \textbf{0.671} &
  {\color[HTML]{FE0000} 0.676} \\
 &
  Mover-1$^{+}$ &
  {\color[HTML]{009901} 0.737} &
  {\color[HTML]{009901} 0.741*} &
  {\color[HTML]{009901} 0.753} &
  {\color[HTML]{009901} 0.733} &
  {\color[HTML]{009901} 0.741} &
  {\color[HTML]{009901} 0.755*} &
  {\color[HTML]{009901} 0.662} &
  {\color[HTML]{009901} 0.685*} &
  {\color[HTML]{009901} 0.700} &
  {\color[HTML]{009901} 0.722*} &
  {\color[HTML]{009901} 0.702*} &
  {\color[HTML]{009901} 0.704*} \\
\multirow{-5}{*}{Reproduced} &
  Bary-W$^{+}$ &
  {\color[HTML]{FE0000} 0.744} &
  {\color[HTML]{FE0000} 0.734} &
  {\color[HTML]{FE0000} 0.771*} &
  {\color[HTML]{FE0000} 0.734*} &
  {\color[HTML]{FE0000} 0.746*} &
  {\color[HTML]{FE0000} 0.751} &
  {\color[HTML]{009901} 0.680*} &
  {\color[HTML]{FE0000} 0.667} &
  {\color[HTML]{009901} 0.701*} &
  {\color[HTML]{FE0000} 0.719} &
  {\color[HTML]{FE0000} 0.702} &
  {\color[HTML]{FE0000} 0.703} \\ \midrule
 &
  BERT-F1 &
  0.735 &
  0.707 &
  0.725 &
  0.705 &
  0.718 &
  0.736 &
  \textbf{0.646} &
  0.646 &
  0.641 &
  0.676 &
  \textbf{0.671} &
  0.669 \\
 &
  Mover &
  0.701 &
  0.694 &
  0.700 &
  0.655 &
  0.688 &
  0.695 &
  0.591 &
  0.628 &
  0.622 &
  0.654 &
  0.640 &
  0.638 \\
 &
  Bary &
  \textbf{0.738} &
  \textbf{0.722} &
  \textbf{0.745} &
  \textbf{0.706} &
  \textbf{0.728} &
  \textbf{0.743} &
  0.642 &
  \textbf{0.664} &
  \textbf{0.664} &
  \textbf{0.714} &
  \textbf{0.671} &
  \textbf{0.683} \\
 &
  Mover$^{+}$ &
  0.711 &
  0.682 &
  0.720 &
  0.647 &
  0.690 &
  0.704 &
  0.607 &
  0.622 &
  0.626 &
  0.660 &
  0.607 &
  0.638 \\
\multirow{-5}{*}{Reported} &
  Bary$^{+}$ &
  0.752* &
  0.745* &
  0.787* &
  0.750* &
  0.759* &
  0.762* &
  0.677* &
  0.683* &
  0.695* &
  0.730 &
  0.705 &
  0.709* \\ \bottomrule
\end{tabular}%
}
\caption{Reproduction: Segment-level Spearman correlation on WMT15-16 using evaluation script provided by \citet{colombo2021automatic}. Reported values are cited from \citet{colombo2021automatic}. $^{+}$ represents using the fine-tuned bert-base-uncased model on MNLI. Values in green/red denote the reproduced results are better/worse than the reported. Bold values refer to the best results with bert-base-uncased model. Values with * denote the best reproduced/reported results.}
\label{tab:repro_wmt1516_spearman}

\centering
\resizebox{\textwidth}{!}{%
\begin{tabular}{@{}c|llllll|lllllll@{}}
\toprule
\multicolumn{1}{l|}{} &
  \multicolumn{6}{c|}{WMT15} &
  \multicolumn{7}{c}{WMT16} \\
\multicolumn{1}{l|}{} &
  metric &
  \multicolumn{1}{c}{cs-en} &
  \multicolumn{1}{c}{de-en} &
  \multicolumn{1}{c}{fi-en} &
  \multicolumn{1}{c}{ru-en} &
  \multicolumn{1}{c|}{avg} &
  \multicolumn{1}{c}{cs-en} &
  \multicolumn{1}{c}{de-en} &
  \multicolumn{1}{c}{ru-en} &
  \multicolumn{1}{c}{fi-en} &
  \multicolumn{1}{c}{ro-en} &
  \multicolumn{1}{c}{tr-en} &
  \multicolumn{1}{c}{avg} \\ \midrule
 &
  BERT-F1 &
  {\color[HTML]{009901} \textbf{0.559*}} &
  {\color[HTML]{009901} \textbf{0.541}} &
  {\color[HTML]{009901} 0.547} &
  {\color[HTML]{009901} \textbf{0.532}} &
  {\color[HTML]{009901} \textbf{0.545}} &
  {\color[HTML]{009901} \textbf{0.564}} &
  {\color[HTML]{009901} 0.474} &
  {\color[HTML]{009901} 0.483} &
  {\color[HTML]{009901} \textbf{0.484}} &
  {\color[HTML]{009901} \textbf{0.520}} &
  {\color[HTML]{FE0000} 0.485} &
  {\color[HTML]{009901} \textbf{0.502}} \\
 &
  Mover-1 &
  {\color[HTML]{009901} 0.537} &
  {\color[HTML]{009901} 0.538} &
  {\color[HTML]{009901} 0.540} &
  {\color[HTML]{009901} 0.515} &
  {\color[HTML]{009901} 0.532} &
  {\color[HTML]{009901} 0.552} &
  {\color[HTML]{009901} 0.458} &
  {\color[HTML]{009901} \textbf{0.488}} &
  {\color[HTML]{009901} 0.475} &
  {\color[HTML]{009901} 0.510} &
  {\color[HTML]{009901} 0.487} &
  {\color[HTML]{009901} 0.495} \\
 &
  Bary-W &
  {\color[HTML]{009901} 0.551} &
  {\color[HTML]{FE0000} 0.528} &
  {\color[HTML]{009901} \textbf{0.566}} &
  {\color[HTML]{FE0000} 0.525} &
  {\color[HTML]{FE0000}0.543} &
  {\color[HTML]{FE0000} 0.543} &
  {\color[HTML]{FE0000} \textbf{0.477}} &
  {\color[HTML]{FE0000} 0.471} &
  {\color[HTML]{FE0000} 0.484} &
  {\color[HTML]{FE0000} 0.513} &
  {\color[HTML]{FE0000} \textbf{0.491}} &
  {\color[HTML]{FE0000} 0.496} \\
 &
  Mover-1$^{+}$ &
  {\color[HTML]{009901} 0.544} &
  {\color[HTML]{009901} 0.556*} &
  {\color[HTML]{009901} 0.569} &
  {\color[HTML]{009901} 0.546*} &
  {\color[HTML]{009901} 0.554} &
  {\color[HTML]{009901} 0.566*} &
  {\color[HTML]{009901} 0.486} &
  {\color[HTML]{009901} 0.499*} &
  {\color[HTML]{009901} 0.513} &
  {\color[HTML]{009901} 0.533*} &
  {\color[HTML]{009901} 0.518*} &
  {\color[HTML]{009901} 0.519*} \\
\multirow{-5}{*}{Reproduced} &
  Bary-W$^{+}$ &
  {\color[HTML]{FE0000} 0.549} &
  {\color[HTML]{FE0000} 0.553} &
  {\color[HTML]{FE0000} 0.580*} &
  {\color[HTML]{FE0000} 0.546*} &
  {\color[HTML]{FE0000} 0.557*} &
  {\color[HTML]{FE0000} 0.561} &
  {\color[HTML]{FE0000} 0.499*} &
  {\color[HTML]{FE0000} 0.485} &
  {\color[HTML]{009901} 0.516*} &
  {\color[HTML]{FE0000} 0.529} &
  {\color[HTML]{FE0000} 0.518} &
  {\color[HTML]{FE0000} 0.518} \\ \midrule
 &
  BERT-F1 &
  0.543 &
  0.529 &
  0.541 &
  0.525 &
  0.535 &
  0.555 &
  0.463 &
  0.469 &
  0.470 &
  0.495 &
  0.490 &
  0.490 \\
 &
  Mover &
  0.520 &
  0.503 &
  0.523 &
  0.469 &
  0.504 &
  0.526 &
  0.442 &
  0.448 &
  0.451 &
  0.482 &
  0.437 &
  0.464 \\
 &
  Bary &
  \textbf{0.549} &
  \textbf{0.531} &
  \textbf{0.563} &
  \textbf{0.532} &
  \textbf{0.544} &
  \textbf{0.563} &
  \textbf{0.479} &
  \textbf{0.481} &
  \textbf{0.483} &
  \textbf{0.529} &
  \textbf{0.514} &
  \textbf{0.508} \\
 &
  Mover$^{+}$ &
  0.520 &
  0.503 &
  0.529 &
  0.473 &
  0.506 &
  0.534 &
  0.448 &
  0.452 &
  0.458 &
  0.486 &
  0.449 &
  0.471 \\
\multirow{-5}{*}{Reported} &
  Bary$^{+}$ &
  0.569* &
  0.562* &
  0.597* &
  0.567* &
  0.574* &
  0.575* &
  0.500* &
  0.513* &
  0.509* &
  0.545* &
  0.524* &
  0.528* \\ \bottomrule
\end{tabular}%
}
\caption{Reproduction: Segment-level Kendall correlation on WMT15-16 using evaluation script provided by \citet{colombo2021automatic}. Reported values are cited from \citet{colombo2021automatic}. $^{+}$ represents using the fine-tuned bert-base-uncased model on MNLI. Values in green/red denote the reproduced results are better/worse than the reported. Bold values refer to the best results with bert-base-uncased model. Values with * denote the best reproduced/reported results.}
\label{tab:repro_wmt1516_kendall}

\end{table*}

\subsection{Reproduction of other tasks}\label{sec:reproduction_other}

\citet{zhao2019moverscore} released the evaluation scripts for WMT17 and TAC2008/2009 and the corresponding datasets on a github\footnote{\url{https://github.com/AIPHES/emnlp19-moverscore}}.  We take them as the resources for reproduction. As for IC and D2T generation evaluation, we write our own evaluation scripts and download those datasets on our own. We obtained MSCOCO, BAGEL, and FSHOTEL datasets from an open question\footnote{\url{https://github.com/AIPHES/emnlp19-moverscore/issues/16}} on its Github page, where \citet{zhao2019moverscore} provided the links to download them. Since \citet{zhao2019moverscore} did not provide much information about how they evaluated on MSCOCO, we also inspect the BERTScore paper \citep{zhang2019bertscore}, where the authors  
gave details of the evaluation process. As each system caption in MSCOCO has multiple references, it is critical to know how to obtain the caption-level scores. \citet{zhang2019bertscore} clearly state that they use the maximal score for each caption as its final score. According to the evaluation scripts for TAC2008/2009 from \citet{zhao2019moverscore}, they averaged the scores for each summary to obtain the summary-level scores, so we assume they might apply the same strategy on MSCOCO. Therefore, we test these two strategies in the reproduction experiment for IC. To check the reliability of our evaluation script, we use it to reproduce the results reported in the BERTScore paper as well. If one can get comparable correlations and the other can not, it may suggest that the authors did extra processing to achieve the results, such as more preprocessing steps on the dataset. The configurations of the evaluation metrics used here are the same as in reproduction attempts on MT. 

\begin{table}[h]

\centering
\resizebox{\columnwidth}{!}{%
\begin{tabular}{@{}lcccccc@{}}
\toprule
dataset                & \multicolumn{3}{c}{BAGEL} & \multicolumn{3}{c}{SFHOTEL} \\
criteria               & Inf     & Nat    & Qual   & Inf     & Nat     & Qual    \\ \midrule
original               & 0.285   & 0.195  & 0.158  & 0.207   & 0.270   & 0.183   \\
reproduced             & 0.244   & 0.145  & 0.092  & \textbf{0.223}   & 0.167   & 0.065   \\
reproduced (stopwords) & 0.230   & 0.135  & 0.078  & \textbf{0.208}   & 0.145   & 0.042   \\ \bottomrule
\end{tabular}%
}
\caption{\emph{Reproduction}: Utterance-level Spearman correlations of MoverScore-1 on BAGEL and SFHOTEL datasets. Original results are citet from \citet{zhao2019moverscore}. Bold values refer to the reproduced resutls that are better than the original.}
\label{tab:dtt reproduction}
\end{table}

\begin{table}[h]

\centering
\resizebox{\columnwidth}{!}{%
\begin{tabular}{@{}lcccc@{}}
\toprule
metric                      & \multicolumn{2}{c}{MoverScore-1} & \multicolumn{2}{c}{BERTScore-R} \\
criteria                    & M1              & M2             & M1             & M2             \\ \midrule
original                    & 0.813           & 0.810          & 0.834          & 0.783          \\
reproduced (mean)           & 0.687            & 0.674           & -              & -              \\
reproduced  (max)           & 0.690           & 0.714          & \textbf{0.851}          & \textbf{0.793}          \\
reproduced (mean+stopwords) & 0.707           & 0.709          & -              & -              \\
reproduced (max+stopwords)  & 0.686           & 0.718          & -              & -              \\ \bottomrule
\end{tabular}%
}
\caption{\emph{Reproduction}: System-level Pearson correlations of MoverScore-1 and BERTScore-R on MSCOCO dataset. Original results are citet from \citet{zhao2019moverscore} and \citet{zhang2019bertscore}. Bold values refer to the reproduced resutls that are better than the original.}
\label{tab:coco reproduction}

\end{table}

\paragraph{Results}
Overall, we could only reconstruct the identical values for summarization in these 3 reproduction attempts. 

Table \ref{tab:dtt reproduction} displays the reproduction results for D2T generation. The reproduced scores with/without stopwords removal go down 0.1/0.08 on average. The maximum deviation is reached in the evaluation of quality on SFHOTEL, down up to 0.14 absolute Spearman correlation. Only two reproduced values are higher than the original, which are the results for informativeness on SFHOTEL dataset. Besides, the reproduced values also deviate least in the assessment of this criterion on both datasets. As for IC, as Table \ref{tab:coco reproduction} shows, the correlations for MoverScore are down by over 0.1 across all evaluation setups.
Nevertheless, BERTScore-Recall performs even on average 0.03 better in our evaluation. This kind of inconsistency between the reproduction results for these two evaluation metrics may suggest that \citet{zhao2019moverscore} did more preprocessing in the evaluation of IC, which is impossible for others to identify if the authors neither document them nor share the relevant code. In contrast, although different preprocessing schemes were applied to MT and summarization evaluation, it is possible to reproduce most of the values because \citet{zhao2019moverscore} released the evaluation scripts. All of the facts mentioned imply the importance of sharing code and data for reproducibility. However, even with the author-provided code and datasets, there is no guarantee that the results can be perfectly reproduced. The authors may ignore some details of the evaluation setup or metric configurations.

\subsection{Subwords, Stopwords, Punctuation}\label{sec:subwords}

\paragraph{Subword Removal} 
BERT leverages a subword-level tokenizer, which breaks a word into subwords when the full word is excluded from its built-in vocabulary (e.g., \emph{smarter $\rightarrow$ smart, \#\#er}). BERT automatically tags all subwords except the first one with \#\#, so we can easily remove them. There are two advantages to doing so. Firstly, it can speed up the system due to the smaller number of embeddings to process. Secondly, it is sometimes equally effective to lemmatization or stemming. E.g., the suffix \emph{er} of the word \emph{smarter} can be removed with this. In some cases, it may keep a less informative part; e.g., 
the prefix \emph{un} in the word \emph{unhappy}.

\paragraph{Stopwords Removal \& Punctuation Removal} 
Both of these two common preprocessing techniques aim to remove 
less relevant 
parts of the text data. A typical stopword list consists of function words such as prepositions  
articles  
and conjunctions. 
As an example, MoverScore achieves a higher correlation with human judgments when removing stopwords on text summarization.

\subsection{Default configuration of evaluation metrics}\label{sec:default}

\begin{itemize}
    \item \textbf{MoverScore} For English evaluation, we use the released version of MoverScore, which makes use of 1) BERT base uncased model finetuned on MNLI dataset, 2) the embeddings of the last five layers aggregated by power means, 3) punctuation removal and the first subword, and 4) IDF-weighting from references and hypotheses separately. we disable stopwords removal in the whole experiment except stopwords tests. For other languages, we replace the model with multilingual BERT base uncased, to keep in line with English evaluation. 
    
    \item \textbf{BERTScore} For English evaluation, we use BERTScore incorporating with BERT base uncased model, the default layer 9, and IDF-weighting from the references. For other languages, similar to MoverScore, we replace the model with multilingual BERT base uncased model.

\end{itemize}

\subsection{Stopword lists}\label{sec:otherlangs}
For English, the first stopword list is obtained from the Github repository of MoverScore\footnote{\url{https://github.com/AIPHES/emnlp19-moverscore/blob/master/examples/stopwords.txt}}, which contains 153 words. 
Since users may first choose existing stopword lists from popular libraries, we consider the stopword lists from NLTK \citep{bird2009natural} and SpaCy \citep{spacy2}, which consist of 179 and 326 words, respectively.

We obtain the stopword lists for other languages from:
\begin{enumerate}[label=\Roman*.]
    \item NLTK \citep{bird2009natural};
    \item SpaCy \citep{spacy2};
    \item a Github repository containing stopword lists for many languages;  \footnote{\url{https://github.com/orgs/stopwords-iso/repositories?type=all}}
    \item a dataset on Kaggle containing stopword lists for many languages \footnote{\url{https://www.kaggle.com/heeraldedhia/stop-words-in-28-languages}};
    \item a Github repository containing Chinese stopword lists \footnote{\url{https://github.com/goto456/stopwords}};
    \item a web containing stopword lists for many languages \footnote{\url{https://countwordsfree.com/stopwords/}}.
\end{enumerate}
Below are the size of each stopword list and its resource:
\begin{itemize}
    \item tr: 551(\Rmnum{2}); 53(\Rmnum{1}); 504(\Rmnum{3});
    \item de: 543(\Rmnum{2}); 231(\Rmnum{1}) 620(\Rmnum{3});
    \item ru: 264(\Rmnum{2}); 151(\Rmnum{1}) 556(\Rmnum{3});
    \item cs: 423(\Rmnum{3}); 405(\Rmnum{6}) 256(\Rmnum{4});
    \item fi: 747(\Rmnum{6}); 847(\Rmnum{3}) 229(\Rmnum{4});
    \item zh: 747(\Rmnum{5}); 1891(\Rmnum{2}) 794(\Rmnum{3});
\end{itemize}

\subsection{Other results for stopwords}\label{sec:stop_best}

\begin{table*}[!htb]
\centering
\resizebox{0.85\textwidth}{!}{%
\begin{tabular}{@{}lcccc|cccc@{}}
\toprule
\multicolumn{5}{c|}{\textbf{Segment-level}} & \multicolumn{4}{c}{\textbf{System-level}} \\
Metric & WMT17-$r$ & WMT18-$\tau$ & WMT19-$\tau$ & AVG & WMT17-$r$ & WMT18-$r$ & WMT19-$r$ & AVG \\ \midrule
Mover-1 & 2.18\% & 2.00\% & 1.42\% & 1.87\% & 0.44\% & 0.20\% & 0.12\% & 0.25\% \\
Mover-2 & 2.09\% & 2.04\% & 1.99\% & 2.04\% & 0.14\% & 0.16\% & 0.20\% & 0.17\% \\
BERT-F1 & 8.74\% & 8.18\% & 6.24\% & 7.72\% & 0.16\% & 0.48\% & 0.25\% & 0.30\% \\ \bottomrule
\end{tabular}%
}
\caption{\texttt{CV}$_{\textbf{STOP}}$ on WMT17-19 to-English language pairs.}
\label{tab:stop_wmt}
\end{table*}
\vspace{1cm}

\begin{figure*}[!h]
    \centering
    \includegraphics[width=0.85\textwidth]{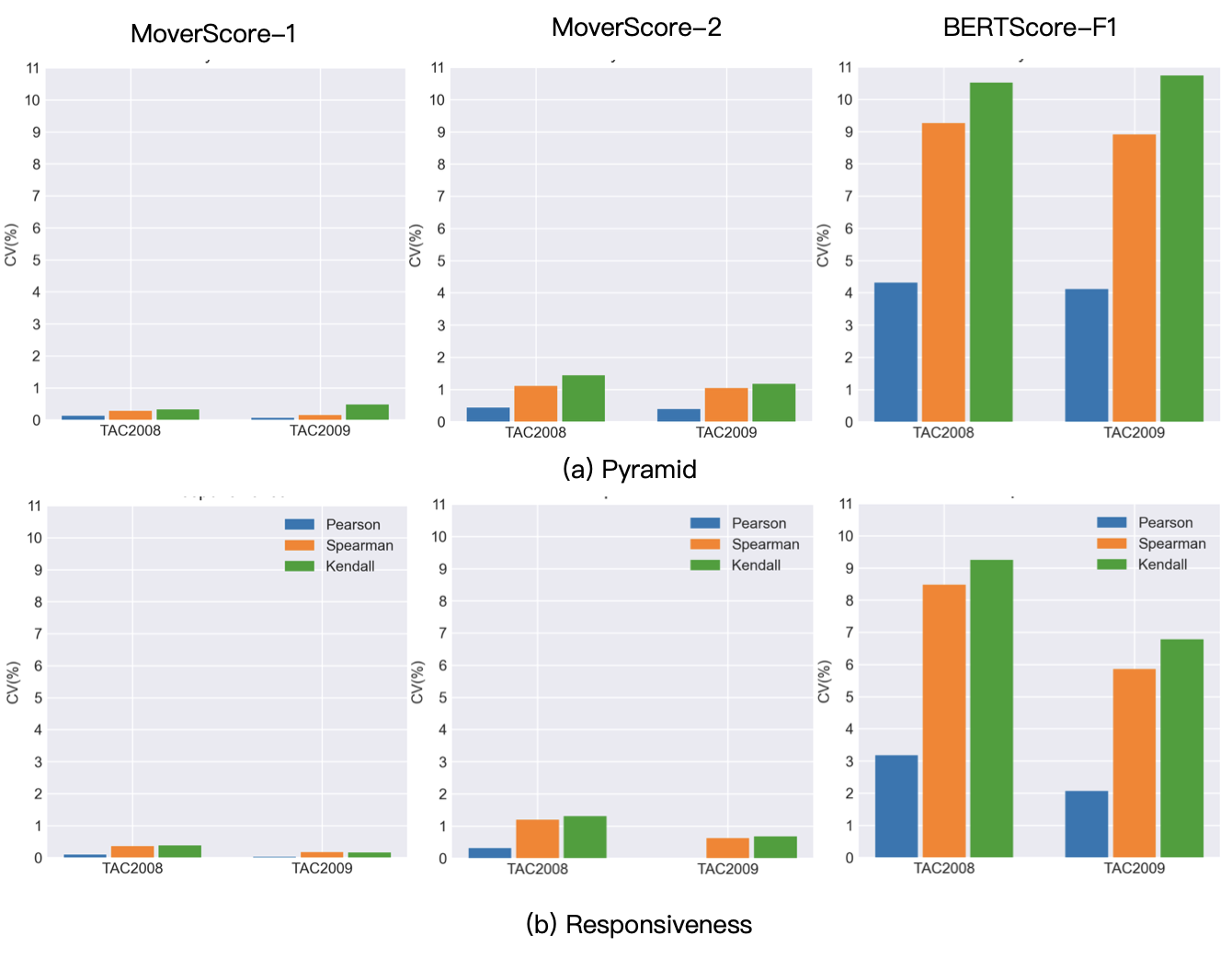}
    \caption{\texttt{CV}$_{\textbf{STOP}}$ on TAC2008-2009.}
    \label{fig:stop_tac}
\end{figure*}

\begin{figure*}[!htb]
    
\scalebox{0.9}{
\begin{minipage}[t]{0.55\textwidth}
\flushleft
\includegraphics[height=0.47\textheight]{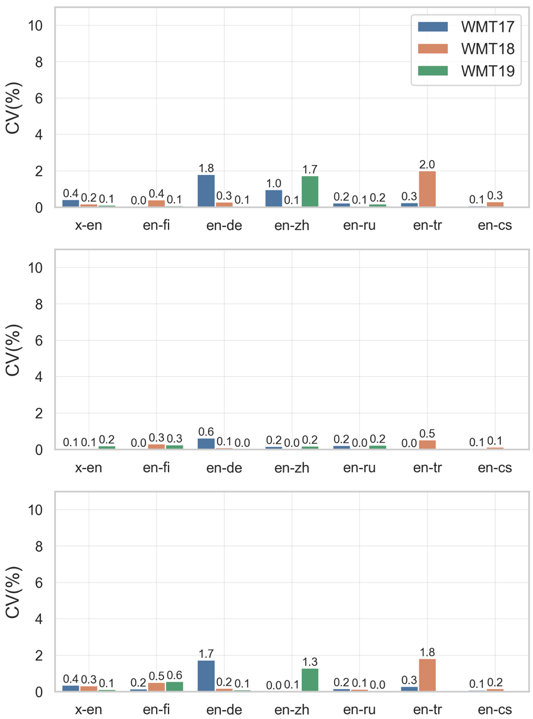}
\caption{From top to bottom: \texttt{CV}$_{\text{STOP}}$, \texttt{CV}$_{\text{IDF}}$, \texttt{CV}$_{\text{SUB}}$. WMT17-19, system-level evaluation, MoverScore-1.}
\label{fig:mover1_cv_sys}
\end{minipage}%
\hspace{0.5cm}
\begin{minipage}[t]{0.38\textwidth}
\centering
\includegraphics[height=0.47\textheight]{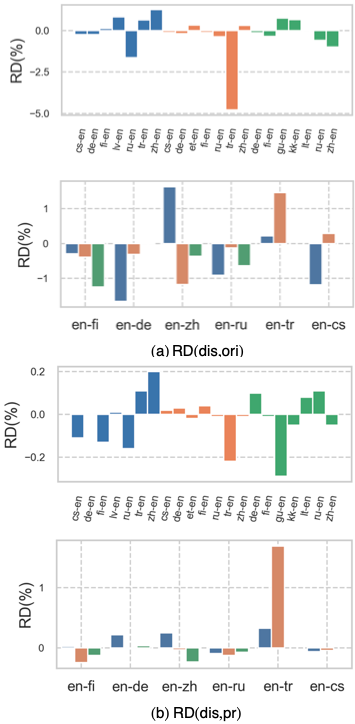}
\centering
\caption{\texttt{RD}(dis,ori), \texttt{RD}(dis,pr). WMT17-19, system-level evaluation, MoverScore-1.}
\label{fig:mover1_rd_sys}
\end{minipage}%
}

\end{figure*}

\begin{figure*}[!htb]
    
\scalebox{0.9}{
\begin{minipage}[t]{0.65\textwidth}
\flushleft
\includegraphics[height=0.22\textheight]{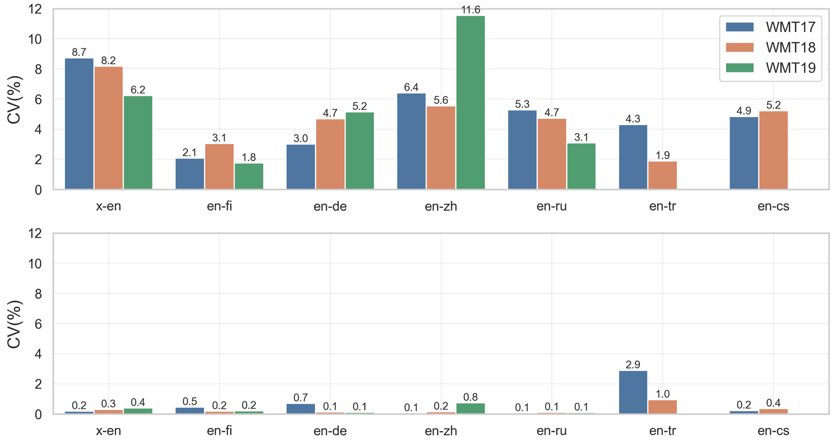}
\caption{From top to bottom: \texttt{CV}$_{\text{STOP}}$, \texttt{CV}$_{\text{IDF}}$. WMT17-19, segment-level evaluation, BERTScore-F1.}
\label{fig:bert_cv_seg}
\end{minipage}%
\hspace{0.5cm}
\begin{minipage}[t]{0.35\textwidth}
\centering
\includegraphics[height=0.22\textheight]{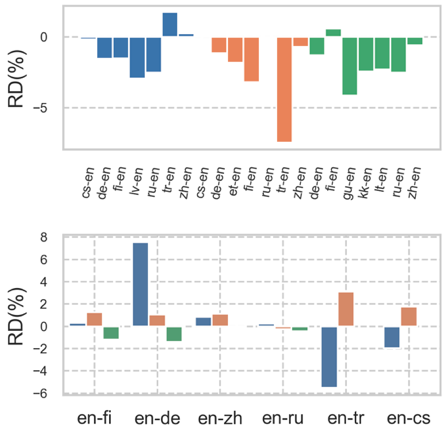}
\centering
\caption{\texttt{RD}(dis,ori). WMT17-19, segment-level evaluation, BERTScore-F1.}
\label{fig:bert_rd_seg}
\end{minipage}%
}

\vspace{3cm}

\scalebox{0.9}{
\begin{minipage}[t]{0.65\textwidth}
\flushleft
\includegraphics[height=0.22\textheight]{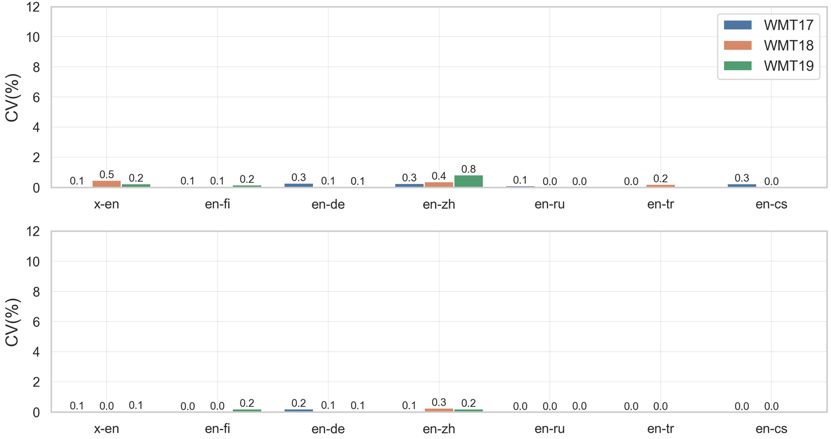}
\caption{From top to bottom: \texttt{CV}$_{\text{STOP}}$, \texttt{CV}$_{\text{IDF}}$. WMT17-19, system-level evaluation, BERTScore-F1.}
\label{fig:bert_cv_sys}
\end{minipage}%
\hspace{0.5cm}
\begin{minipage}[t]{0.35\textwidth}
\centering
\includegraphics[height=0.22\textheight]{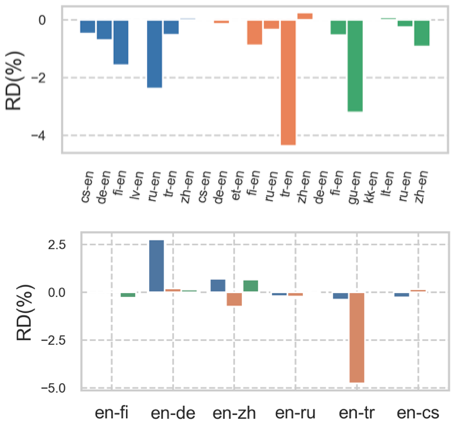}
\centering
\caption{\texttt{RD}(dis,ori). WMT17-19, system-level evaluation, BERTScore-F1.}
\label{fig:bert_rd_sys}
\end{minipage}%
}

\end{figure*}

Table \ref{tab:stop_wmt} and Figure \ref{fig:stop_tac} display the \texttt{CV}$_{\text{STOP}}$ in English evaluation. We can observe that: (i) Among the three evaluation metrics, MoverScore-1 is least sensitive to stopwords removal, while BERTScore-F1 behaves most sensitively. (ii) The metrics are most sensitive in segment-level MT evaluation among the examined evaluation tasks. (iii) Kendall's $\tau$ varies most with changing stopword settings, while Pearson is least sensitive. In other language environment, we can also observe that the metrics are more sensitive at segment-level than at system-level (Figure \ref{fig:mover1_cv_sys}, \ref{fig:bert_cv_seg} and \ref{fig:bert_cv_sys} (top)). Further, except for Chinese and English, where BERTScore-F1 behaves much more sensitively than MoverScore-F1, the difference between their sensitivity is less pronounced (see Figure \ref{fig:stopwords} and \ref{fig:bert_cv_seg} (top)).

\begin{figure*}[!htb]
\centering
\includegraphics[height=0.38\textheight
]{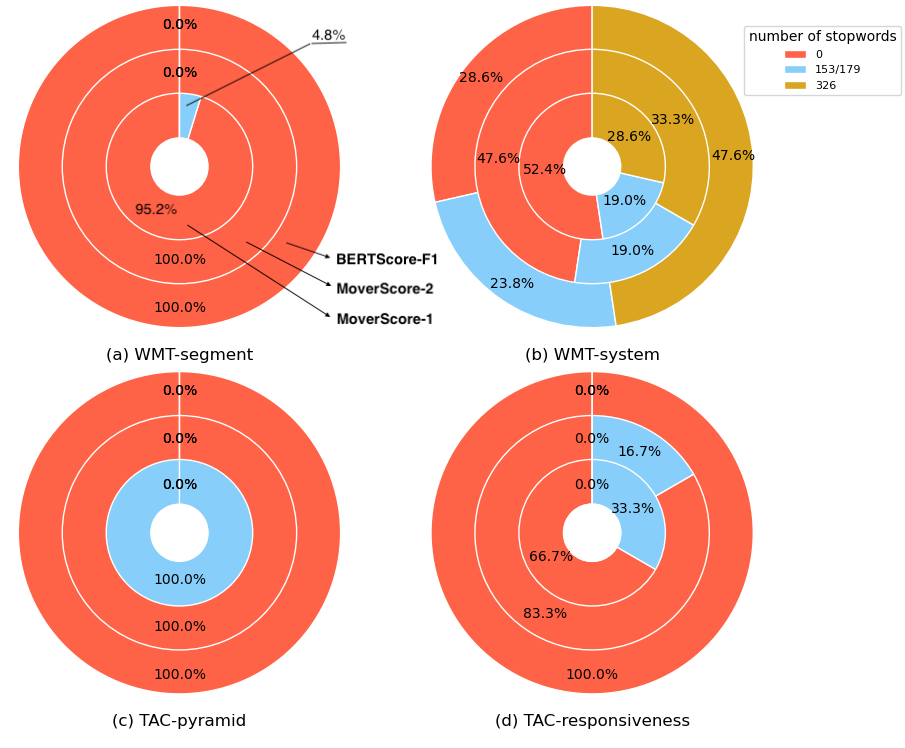}
\caption{
Distribution of the best stopword setting of each evaluation metric on each evaluation task for \textbf{English}. The rings from the inside to the outside represent MoverScore-1, MoverScore-2 and BERTScore-F1. For MT, each language pair in WMT datasets is regarded as a test case, resulting in 21 test cases (3 datasets times 7 language pairs). For summarization tasks, each type of correlation is regarded as a test case for each criterion, resulting in 6 test cases (3 correlations times 2 datasets). The MoverScore (153) and SpaCy (179) stopword lists yield exactly the same results.}

\label{fig:sw_distribution_en}

\end{figure*}

\begin{table}[htb]
    \centering
\resizebox{\columnwidth}{!}{%
\begin{tabular}{@{}cccc|ccc@{}}
\toprule
metric  & \multicolumn{3}{c|}{BERTScore-F1} & \multicolumn{3}{c}{MoverScore-1} \\
dataset & WMT17     & WMT18     & WMT19     & WMT17     & WMT18     & WMT19    \\ \midrule
en      & 0         & 0         & 0         & 0         & 0         & 0        \\
zh      & 0         & 0         & 0         & 0         & 0         & 0        \\
de      & 0         & 0         & 0         & 0         & 0         & 0        \\
ru      & 0         & 0         & 0         & 0         & 0         & 0        \\
fi      & 0         & 0         & 0         & 0         & 0         & 229      \\
cs      & 0         & 0         & -         & 0         & 0         & -        \\
tr      & 504       & 551       & -         & 504       & 551       & -        \\ \bottomrule
\end{tabular}%
}
\caption{
Distribution of the best stopword settings for \textbf{all tested languages} in \textbf{segment-level MT} evaluation. Values indicate the size of the stopword lists.}
\label{tab:sw_distribution_all}
\end{table}

Figure \ref{fig:sw_distribution_en} illustrates the distribution of the best stopword settings for \textbf{English}. 
In \textbf{segment-level} MT evaluation (Figure \ref{fig:sw_distribution_en}(a)), there is only one case that the best result is achieved by removing stopwords, which takes place on MoverScore-1.
In contrast, the best stopword lists for \textbf{system-level MT} evaluation can be any of the settings for all evaluation metrics (Figure \ref{fig:sw_distribution_en}(b)). However, in about 50\% of the test cases, MoverScore still performs best when disabling stopwords removal. In \textbf{Pyramid} evaluation (Figure \ref{fig:sw_distribution_en}(c)), MoverScore-1 achieves the best results using the original stopword list for all test cases, whereas disabling stopwords removal is still the best choice for MoverScore-2 and BERTScore-F1. In the evaluation of \textbf{Responsiveness} ((Figure \ref{fig:sw_distribution_en}(d))), two cases (33.3\%) can be seen that MoverScore-1 applying the original stopword list performs best; this happens only once on MoverScore-2 (16.7\%). BERTScore-F1 never benefit from stopwords removal on all evaluation tasks. 

Further, in Table \ref{tab:sw_distribution_all}, we present the best stopword setting for \textbf{all examined languages} in \textbf{segment-level MT} evaluations. Except Finnish and Turkish, disabling stopwords removal is always the best choice for all other languages. For Finnish, only on one dataset, MoverScore-1 performs better using stopwords removal, whereas, for Turkish, both evaluation metrics achieve the best performance applying the same stopword lists. The reason might be that both Turkish and Finnish belong to agglutinative languages, and those languages tend to have a high rate of affixes or morphemes per word, which means there may exist more noise in word embeddings.

\begin{table*}[!hbt]
\centering
\resizebox{\textwidth}{!}{%
\begin{tabular}{@{}lcccc|cccc@{}}
\toprule
        & \multicolumn{4}{c|}{\textbf{Segment-level}}      & \multicolumn{4}{c}{\textbf{System-level}}  \\
Metric  & WMT17-$r$ & WMT18-$\tau$ & WMT19-$\tau$ & AVG    & WMT17-$r$ & WMT18-$r$ & WMT19-$r$ & AVG    \\ \midrule
Mover-1 & 0.13\%    & 0.67\%       & 2.56\%       & 1.12\% & 0.05\%    & 0.06\%    & 0.21\%    & 0.11\% \\
Mover-2 & 0.78\%    & 1.19\%       & 3.84\%       & 1.94\% & 0.25\%    & 0.17\%    & 0.49\%    & 0.30\% \\
BERT-F1 & 0.20\%    & 0.32\%       & 0.41\%       & 0.31\% & 0.05\%    & 0.02\%    & 0.08\%    & 0.05\% \\ \bottomrule
\end{tabular}%
}
\caption{\texttt{CV}$_{\text{IDF}}$ for WMT17-19 to-English language pairs.}
\label{tab:wmt_idf_cv}

\vspace{15pt}

\resizebox{\textwidth}{!}{%
\begin{tabular}{@{}lcccccc|cccccc@{}}
\toprule
 & \multicolumn{6}{c|}{\textbf{Pyramid}} & \multicolumn{6}{c}{\textbf{Responsiveness}} \\
 & \multicolumn{3}{c}{{\ul TAC2008}} & \multicolumn{3}{c|}{{\ul TAC2009}} & \multicolumn{3}{c}{{\ul TAC2008}} & \multicolumn{3}{c}{{\ul TAC2009}} \\
Metric & $r$ & $\rho$ & $\tau$ & $r$ & $\rho$ & $\tau$ & $r$ & $\rho$ & $\tau$ & $r$ & $\rho$ & $\tau$ \\ \midrule
Mover-1 & 0.11\% & 0.15\% & \multicolumn{1}{c|}{0.40\%} & 0.08\% & 0.04\% & 0.43\% & 0.13\% & 0.27\% & \multicolumn{1}{c|}{0.31\%} & 0.14\% & 0.33\% & 0.37\% \\
Mover-2 & 0.13\% & 0.23\% & \multicolumn{1}{c|}{0.27\%} & 0.11\% & 0.31\% & 0.35\% & 0.19\% & 0.37\% & \multicolumn{1}{c|}{0.39\%} & 0.13\% & 0.42\% & 0.46\% \\
BERT-F1 & 0.19\% & 0.42\% & \multicolumn{1}{c|}{0.51\%} & 0.06\% & 0.24\% & 0.34\% & 0.21\% & 0.32\% & \multicolumn{1}{c|}{0.31\%} & 0.17\% & 0.21\% & 0.19\% \\ \bottomrule
\end{tabular}%
}
\caption{\texttt{CV}$_{\text{IDF}}$ for TAC2008-2009.}
\label{tab:tac_idf_cv}

\end{table*}

\begin{table}[!htb]
\centering
\resizebox{\columnwidth}{!}{%
\begin{tabular}{@{}lcc@{}}
\toprule
Corpora & WMT18-AVG & WMT19-AVG \\
ORI & \textbf{0.355} & \textbf{0.333} \\ \midrule
Wili\_2008(117500) & 0.349 & 0.323 \\
Wikipedia(100000) & 0.350 & 0.320 \\
Wikipedia(1000000) & 0.351 & 0.320 \\
Wikipedia(2500000) & 0.351 & 0.320 \\
Wikipedia(5000000) & 0.350 & 0.320 \\
Wikipedia(7500000) & 0.351 & 0.320 \\
Wikipedia(10000000) & 0.351 & 0.320 \\
IMDB\_train(25000) & 0.347 & 0.323 \\
Wikitext(23767) & 0.350 & 0.324 \\
Wiki40b(2926536) & 0.347 & 0.324 \\ \bottomrule
\end{tabular}%
}
\caption{Average segment-level Kendall correlation of MoverScore-1 using idf$_{\text{large}}$ with human judgements in WMT18-19 to-English language pairs. Bold values refer to the best results. Number in bracket represents the number of documents in this corpus.}
\label{tab:external_en}
\end{table}

\subsection{IDF Corpora}\label{sec:idf_corpora}

\begin{itemize}
    \item \textbf{Wikipedia}\footnote{\url{https://huggingface.co/datasets/wikipedia}} \citep{wikidump} 
    Wikipedia dataset contains clean full articles of Wikipedia pages but with many non-content segments such as citations, links and so on. Due to memory limit, we can only test a few segments in this dataset.
    
    \item \textbf{Wiki40b}\footnote{\url{https://huggingface.co/datasets/wiki40b}} \citep{guo-etal-2020-wiki} This dataset aims at entity identification task, and is cleaned up by excluding ambiguation and non-entity pages from Wikipedia, and non-content and structured part from each page.

    \item \textbf{WikiText}\footnote{\url{https://huggingface.co/datasets/wikitext\#wikitext-2-raw-v1-1}} \citep{merity2016pointer} This is a language modelling dataset, containing texts extracted from the set of verified good and featured articles on English Wikipedia.
    
    \item \textbf{Wili\_2008}\footnote{\url{https://huggingface.co/datasets/wili_2018}} \citep{thoma_martin_2018_841984} The goal of this dataset is to train and test language identification models. It contains short paragraphs of many languages from Wikipedia.
    
    \item \textbf{IMDB}\footnote{\url{https://huggingface.co/datasets/imdb}} \citep{maas-EtAl:2011:ACL-HLT2011} This dataset contains movie reviews and their sentiment label, aiming at binary sentiment classification for English data.

\end{itemize}

\subsection{Other results for IDF-weighting}\label{sec:idf}

\begin{figure*}[hb]
    \centering
    \subfigure[Segment-level]{
    \includegraphics[width=0.8\textwidth]{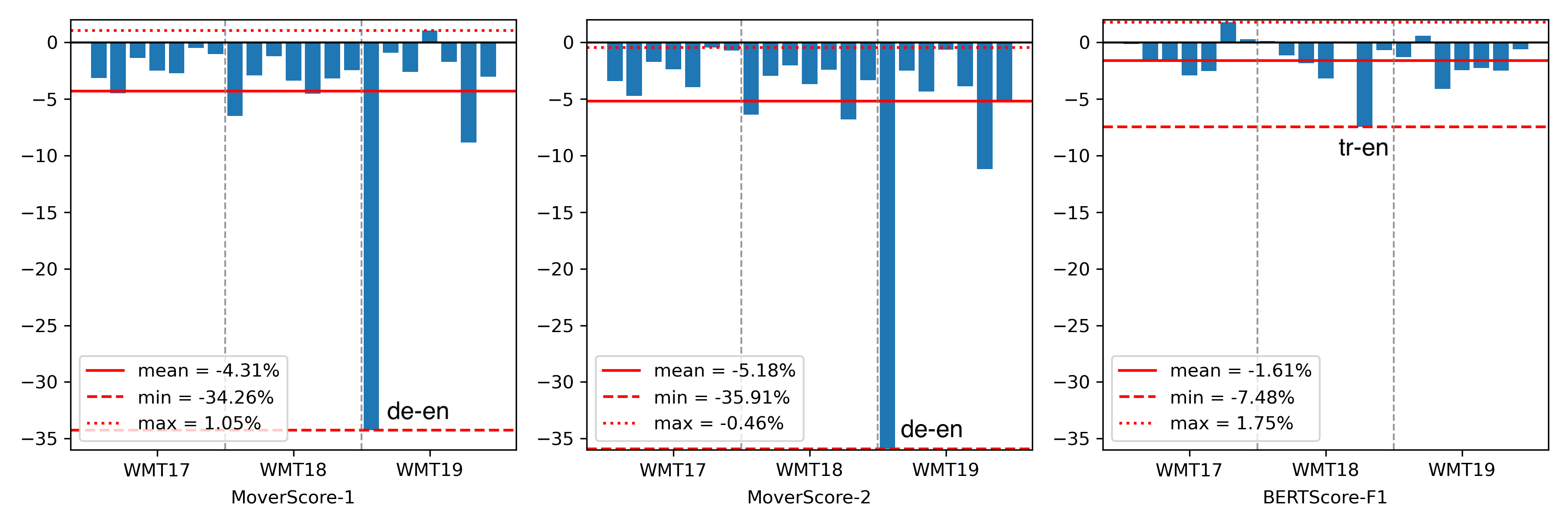}
    }
    \subfigure[System-level]{
    \includegraphics[width=0.8\textwidth]{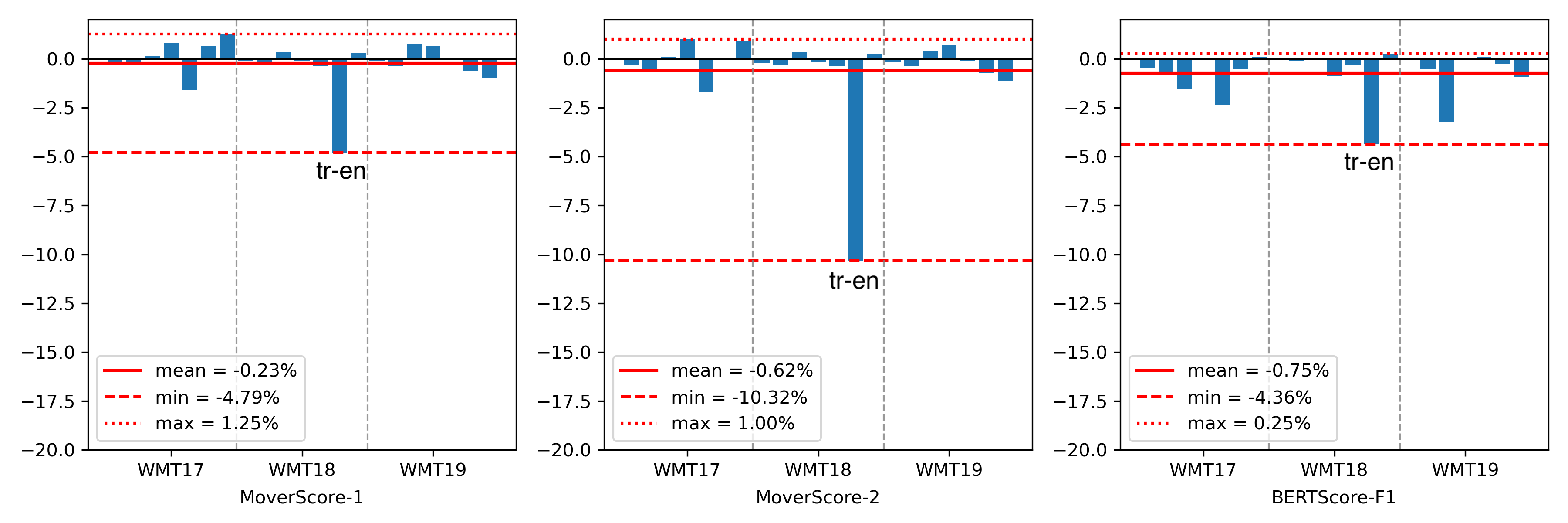}
    }
    \caption{\texttt{RD}(dis,ori), WMT17-19, MT evaluation, MoverScore-1/2 and BERTScore-F1. Negative values indicate idf$_{\text{ori}}$ works better.}
    \label{fig:rd_dis_ori_wmt}
    
    \centering
    \includegraphics[width=0.8\textwidth]{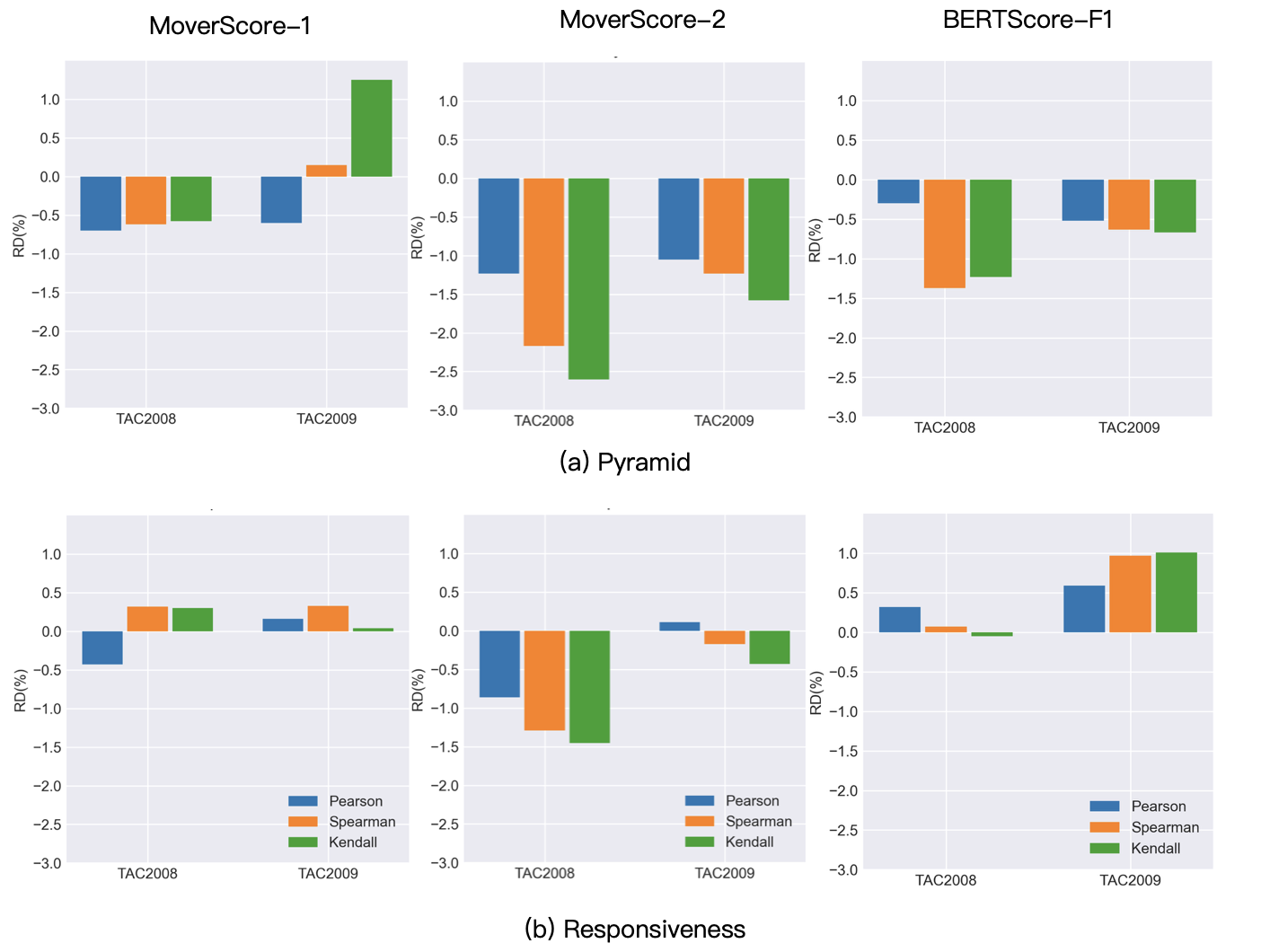}
    \caption{\texttt{RD}(dis,ori), TAC2008-2009, summary-level summarization evaluation, MoverScore-1/2 and BERTScore-F1. Negative values indicate idf$_{\text{ori}}$ works better.}
    \label{fig:rd_idf_sum}

\end{figure*}

\begin{figure*}[hb]
\centering
\subfigure[MoverScore-1]{

\includegraphics[width=\textwidth]{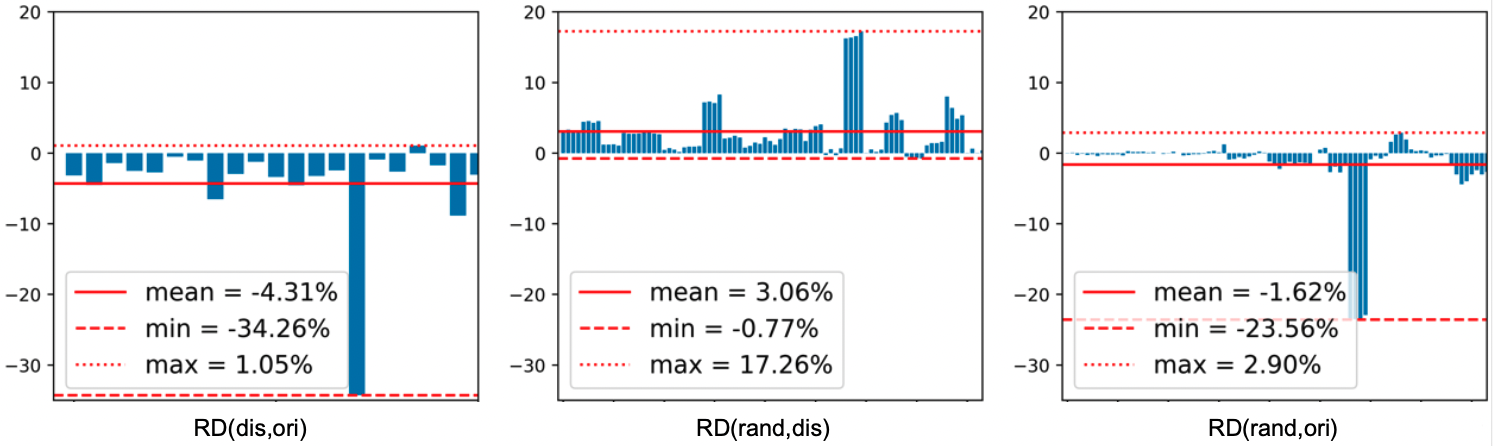}

}
\subfigure[MoverScore-2]{

\includegraphics[width=\textwidth]{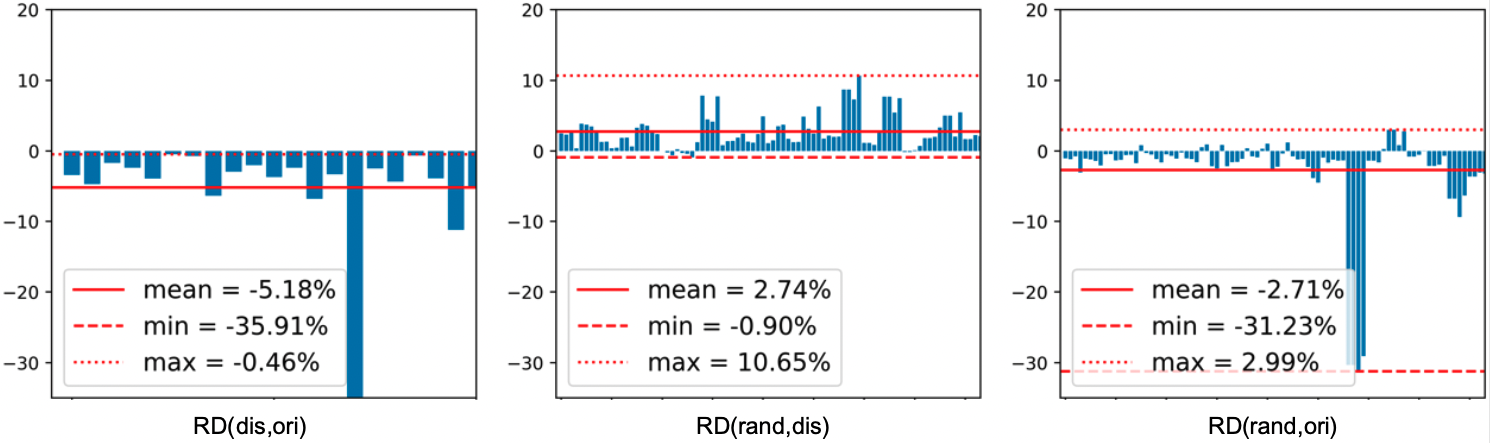}

}
\subfigure[BERTScore-F1]{

\includegraphics[width=\textwidth]{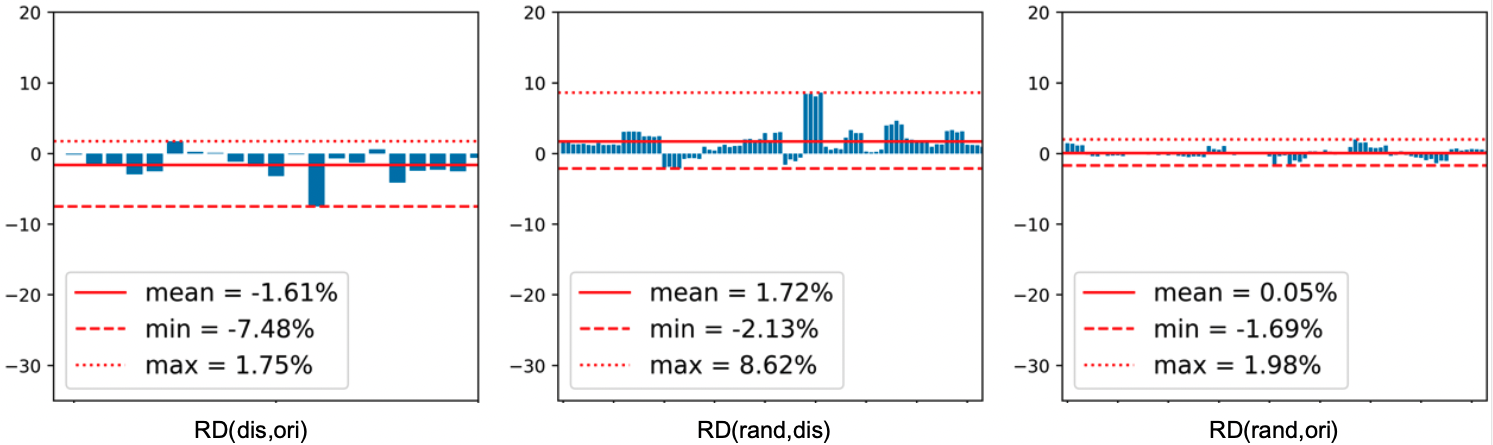}

}
\caption{\texttt{RD}(dis,ori), \texttt{RD}(rand,dis), and \texttt{RD}(rand,ori); WMT17-19, segment-level MT evaluation, MoverScore-1/2 and BERTScore-F1.
Negative values indicate the latter idf$_{\text{x}}$ works better.
}
\label{fig:idf_rd_wmt_seg}
\end{figure*}

As shown in Figure \ref{fig:rd_dis_ori_wmt} and \ref{fig:rd_idf_sum}, in English evaluation, the metric performance of the three evaluation metrics drop most from disabling IDF-weighting in segment-level MT evaluation, where the varying IDF corpora also have the largest impact among the examined evaluation tasks (see Table \ref{tab:wmt_idf_cv} and \ref{tab:tac_idf_cv}). Among the three metrics, BERTScore-F1 is least sensitive to IDF-weighting, to which idf$_{\text{ori}}$ and idf$_{\text{rand}}$ are almost equally effective, whereas idf$_{\text{ori}}$ yields considerably better results than idf$_{\text{rand}}$ for MoverScore-1/2; MoverScore-2 behaves slightly more sensitively than MoverScore-1 (see Figure \ref{fig:idf_rd_wmt_seg}). Moreover, unlike in English evaluation, the contribution of IDF-weighting seems less stable for other languages (see Figure \ref{fig:mover1_rd_seg}(a) and \ref{fig:bert_rd_seg}).

Further, 
Table \ref{tab:external_en} presents the results for idf$_{\text{large}}$ 
in English evaluation. First, the size of those corpora is much larger than the original corpora, but MoverScore still performs better with original IDF-weighting. Secondly, the results for Wikipedia shows that the metric performance does not enhance with the increasing size of IDF corpora. Thirdly, although those corpora contain articles in many domains, they do not provide more applicable IDF-weighting neither. In conclusion, no IDF-weighting from large-domain and large-scale corpora works as well as the original IDF-weighting in segment-level MT evaluation for English, where MoverScore-1 behaves most sensitively to IDF.

\clearpage
\subsection{Best settings of subword selection + PR}\label{sec:sub_best}

\begin{table*}[hb]
\centering
\resizebox{0.75\textwidth}{!}{%
\begin{tabular}{|l|cccccc|cccccc|cccc|}
\hline
\multicolumn{1}{|c|}{} & \multicolumn{6}{c|}{WMT17} & \multicolumn{6}{c|}{WMT18} & \multicolumn{4}{c|}{WMT19} \\ \hline
\multicolumn{1}{|c|}{} & \multicolumn{1}{c|}{de} & \multicolumn{1}{c|}{zh} & \multicolumn{1}{c|}{ru} & \multicolumn{1}{c|}{fi} & \multicolumn{1}{c|}{tr} & cs & \multicolumn{1}{c|}{de} & \multicolumn{1}{c|}{zh} & \multicolumn{1}{c|}{ru} & \multicolumn{1}{c|}{fi} & \multicolumn{1}{c|}{tr} & cs & \multicolumn{1}{c|}{de} & \multicolumn{1}{c|}{zh} & \multicolumn{1}{c|}{ru} & fi \\ \hline
first & \multicolumn{1}{c|}{\textbf{}} & \multicolumn{1}{c|}{} & \multicolumn{1}{c|}{} & \multicolumn{1}{c|}{} & \multicolumn{1}{c|}{} &  & \multicolumn{1}{c|}{\Checkmark} & \multicolumn{1}{c|}{} & \multicolumn{1}{c|}{} & \multicolumn{1}{c|}{} & \multicolumn{1}{c|}{} &  & \multicolumn{1}{c|}{} & \multicolumn{1}{c|}{} & \multicolumn{1}{c|}{} &  \\ \hline
all & \multicolumn{1}{c|}{{\ul }} & \multicolumn{1}{c|}{\Checkmark} & \multicolumn{1}{c|}{} & \multicolumn{1}{c|}{\Checkmark} & \multicolumn{1}{c|}{\Checkmark} & \Checkmark & \multicolumn{1}{c|}{{\ul }} & \multicolumn{1}{c|}{} & \multicolumn{1}{c|}{\Checkmark} & \multicolumn{1}{c|}{{\ul }} & \multicolumn{1}{c|}{} &  & \multicolumn{1}{c|}{} & \multicolumn{1}{c|}{} & \multicolumn{1}{c|}{} &  \\ \hline
ave-all & \multicolumn{1}{c|}{\Checkmark} & \multicolumn{1}{c|}{} & \multicolumn{1}{c|}{} & \multicolumn{1}{c|}{} & \multicolumn{1}{c|}{} &  & \multicolumn{1}{c|}{} & \multicolumn{1}{c|}{\Checkmark} & \multicolumn{1}{c|}{} & \multicolumn{1}{c|}{\Checkmark} & \multicolumn{1}{c|}{} &  & \multicolumn{1}{c|}{} & \multicolumn{1}{c|}{} & \multicolumn{1}{c|}{} & \Checkmark \\ \hline
first+PR* & \multicolumn{1}{c|}{} & \multicolumn{1}{c|}{} & \multicolumn{1}{c|}{} & \multicolumn{1}{c|}{} & \multicolumn{1}{c|}{} &  & \multicolumn{1}{c|}{} & \multicolumn{1}{c|}{} & \multicolumn{1}{c|}{} & \multicolumn{1}{c|}{} & \multicolumn{1}{c|}{} &  & \multicolumn{1}{c|}{} & \multicolumn{1}{c|}{\Checkmark} & \multicolumn{1}{c|}{\Checkmark} &  \\ \hline
all+PR & \multicolumn{1}{c|}{} & \multicolumn{1}{c|}{} & \multicolumn{1}{c|}{\Checkmark} & \multicolumn{1}{c|}{} & \multicolumn{1}{c|}{} &  & \multicolumn{1}{c|}{} & \multicolumn{1}{c|}{} & \multicolumn{1}{c|}{} & \multicolumn{1}{c|}{} & \multicolumn{1}{c|}{\Checkmark} & \Checkmark & \multicolumn{1}{c|}{\Checkmark} & \multicolumn{1}{c|}{} & \multicolumn{1}{c|}{} &  \\ \hline
ave-all+PR & \multicolumn{1}{c|}{} & \multicolumn{1}{c|}{} & \multicolumn{1}{c|}{} & \multicolumn{1}{c|}{} & \multicolumn{1}{c|}{} &  & \multicolumn{1}{c|}{} & \multicolumn{1}{c|}{} & \multicolumn{1}{c|}{} & \multicolumn{1}{c|}{} & \multicolumn{1}{c|}{} &  & \multicolumn{1}{c|}{} & \multicolumn{1}{c|}{} & \multicolumn{1}{c|}{} &  \\ \hline
\end{tabular}%
}
\caption{
Best configuration of MoverScore-1 regarding subwords and punctuations for \textbf{other languages}. WMT17-19, \textbf{segment-level MT} evaluation. We mark the default configuration of MoverScore with $*$.}
\label{tab:sub_best_en}

\vspace{1.2cm}

\centering
\includegraphics[width=0.8\textwidth]{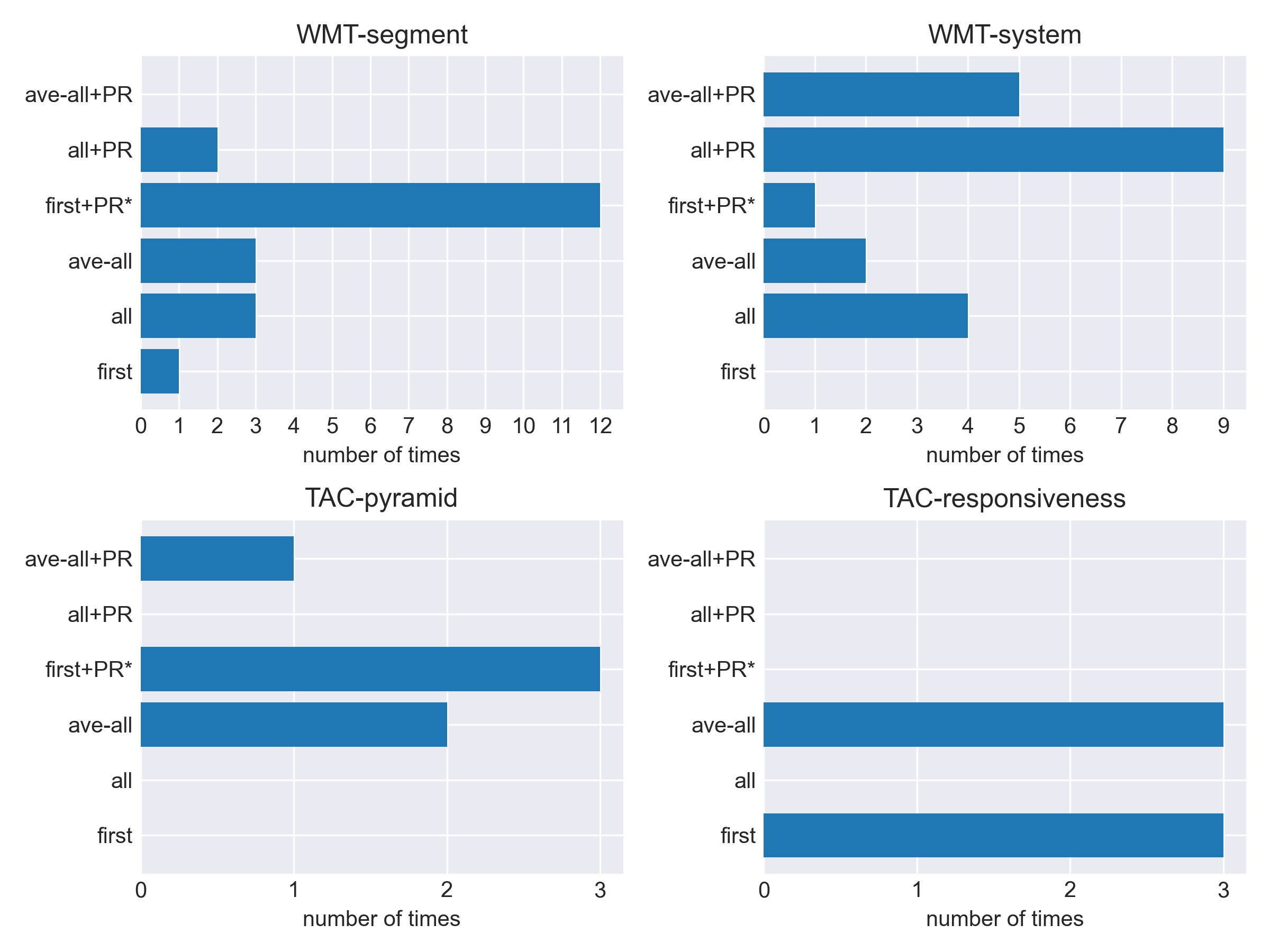}
\caption{
Best configuration of MoverScore-1 regarding subwords and punctuations for \textbf{English}. WMT17-19 and TAC2008-2009. We mark the default configuration of MoverScore with $*$.}
\label{fig:sub_best_other}

\end{table*}

\end{document}